\documentclass[11pt]{article}

\usepackage[final]{acl}

\usepackage{times}
\usepackage{latexsym}

\usepackage[T1]{fontenc}

\usepackage[utf8]{inputenc}

\usepackage{microtype}

\usepackage{inconsolata}

\usepackage{graphicx}

\usepackage{amsmath}
\usepackage{amssymb}
\usepackage{tabularx}
\usepackage{wrapfig}

\usepackage[table]{xcolor}
\usepackage{colortbl}
\usepackage{pgf}
\usepackage{placeins}

\usepackage[skins,breakable]{tcolorbox}
\tcbset{
  promptbox/.style={
    enhanced,
    breakable,
    colback=white,
    colframe=black,
    sharp corners,
    boxrule=0.8pt,
    left=6pt,
    right=6pt,
    top=6pt,
    bottom=6pt,
  }
}

\usepackage{CJKutf8}

\usepackage{subcaption}   

\usepackage{booktabs}     
\usepackage{multirow}
\usepackage{siunitx}      
\usepackage{csvsimple}
\usepackage{soul,todonotes} 
	\definecolor{kmycolor}{rgb}{0.858, 0.188, 0.478}

\title{When Does Mixing Help? Analyzing Query Embedding Interpolation \\ in Multilingual Dense Retrieval }

\author{
  Tongyao Zhu\thanks{Equal contribution} \qquad Chao-Ming Huang$^*$ \qquad Min-Yen Kan\thanks{Corresponding author} \\
  National University of Singapore \\
  \texttt{\{tongyao.zhu, hcm\}@u.nus.edu} \qquad \texttt{knmnyn@nus.edu.sg}
}
\begin{document}
\begin{CJK}{UTF8}{gbsn}
\maketitle

\begin{abstract}
While mixed-language querying is ubiquitous in multilingual communities, the sensitivity of dense retrievers to such queries remains poorly understood. We present a ratio-controlled study on mMARCO that systematically evaluates retrieval performance by varying the mixing proportion of parallel query translations via \emph{embedding-level mixing}---constructing mixed queries as an interpolation of monolingual embeddings. Experiments with BGE-M3 demonstrate that an optimal mixing ratio outperforms the best monolingual endpoint in 88/105 cases. We uncover a distinct asymmetry driven by English dominance: mixing is uniformly beneficial when retrieving from non-English document indices, whereas indices containing English are best served by pure English queries. Furthermore, English acts as the strongest mixing partner for every non-English document language. Finally, when controlling for English dominance, mixing gains correlate negatively with typological distance. We conclude that language-mix sensitivity is structured and predictable, and we validate the robustness of these patterns across model families and scales.\footnote{The code can be found at \url{https://github.com/tongyao-zhu/query-embedding-mix}}

\end{abstract}

\section{Introduction}

Bilingual and multilingual users frequently express a single information need in multiple languages within the same query \citep{Fu-2019-mixed-query}.
Such \emph{code-mixing} (or code-switching)---the blending of elements from multiple languages in a single sentence---is common in search and conversational systems \citep{sitaram2020surveycodeswitchedspeechlanguage}. Recent work has also started to benchmark retrieval under mixed-language queries in bilingual web search~\citep{kim2025milqbenchmarkingirmodels}.
However, multilingual retrievers are typically developed and evaluated under the assumptions that queries are in a single language (either monolingual retrieval, or cross-lingual retrieval), which means mixed-query behavior is largely untested in standard benchmarks \citep{zhang-etal-2021-mr, zhang-etal-2023-miracl, bonifacio2022mmarcomultilingualversionms}. As a result, existing multilingual information retrieval (IR) evaluations provide limited insight into how dense retrievers behave when the query itself mixes languages, and whether mixing can help when the documents to be retrieved involve more than one language.
\begin{figure}[t]
  \centering
  \includegraphics[width=\columnwidth]{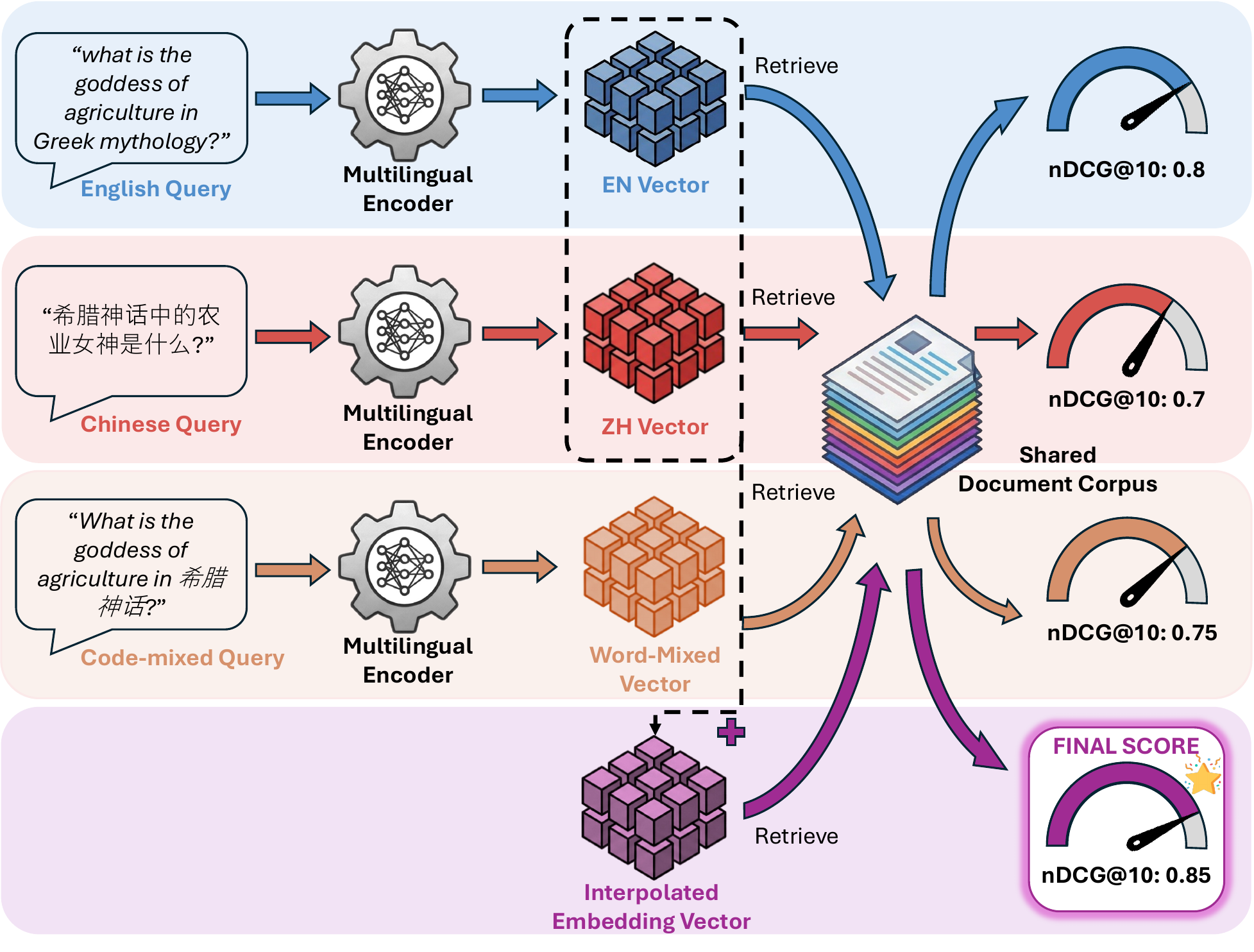}
  \caption{An illustration of the protocol of our study. We use the interpolation of the monolingual query embeddings to serve as the language-mixed representation.} 
  \vspace{-3mm}
  \label{fig:teaser}

\end{figure}
A real question remains: when does language mixing actually help? Specifically, when we vary the mixing ratio between two languages, is retrieval effectiveness bounded by the better monolingual endpoint, or can interior mixtures outperform both?
Answering this requires mixing-ratio control, parallel queries, and evaluation under different document-language compositions.

However, generating high-quality code-mixed queries can be highly resource-intensive \citep{kim2025milqbenchmarkingirmodels}, let alone controlling the mixing ratio. In our preliminary study, we instruct a large language model (LLM) to perform word-level code switching, replacing certain words in the query with another language. While the model is able to generate code-mixed queries, successful generation is not guaranteed, and each mixing ratio requires a separate LLM inference call.
 To avoid these issues, we introduce a novel and simple way of mixing: \emph{embedding-level mixing}.
Given monolingual query embeddings of two languages, we form a convex combination\footnote{We follow the mathematical definition: convex combination is a linear combination of points where all coefficients are non-negative and sum to 1. Similarly, we use the word ``interior'' to describe this type of convex combination.} of the embeddings at different target ratios, and use this interpolation of embeddings as the language-mixed query representation. This enables controlled comparisons across many language pairs unfamiliar to LLMs, and precise control of the mixing ratio without having to trigger LLM generation multiple times under different ratios. We demonstrate that such embedding-level mixing exhibits a similar trend to, and overall higher performance than, word-level mixing by LLM generation, and decide to use it for precise controlled studies on the mixing behavior.

We next study the behavior of a multilingual dense retriever, BGE-M3~\citep{chen2024bgem3}, with language-mixed queries. We use parallel query translations from mMARCO~\citep{bonifacio2022mmarcomultilingualversionms}. We study 35 language pairs, and for each pair $(L_1, L_2)$, we evaluate three document--language settings: $L_1$-only, $L_2$-only, and bilingual union of $L_1{+}L_2$ documents. These settings mirror realistic setups where the relevant information may be written in only one language or prevalent across many languages through translation.

Our results show that embedding-mixing effects are systematic and depend heavily on the query and document languages: 
\textbf{(1)~Mixing often helps:}~Optimal embedding-level mixing ratios outperform the best monolingual queries in 88 of 105 (83.8\%) settings. 
\textbf{(2)~English asymmetry:} Mixing is uniformly beneficial when English is absent from the index; however, indices containing English documents are best served by pure English queries, revealing an asymmetric dominance of English in multilingual vector spaces. 
\textbf{(3)~Strongest partner:}~English acts as the strongest mixing partner for every non-English document language. 
\textbf{(4)~Language metadata:}~When controlling for English dominance, mixing gains decrease as the typological distance between query languages increases. Other metadata (e.g. scripts and resource level) only show weak correlations. 
\textbf{(5)~Robustness:}~These patterns generally hold across model families and scales, suggesting embedding-mix sensitivity is a predictable property of current retrievers. 
Together, these results characterize when and why mixing two languages helps multilingual dense retrieval and provide actionable guidance for multilingual IR and retrieval-augmented pipelines that must operate on mixed-language user inputs.

\section{Related Work}
\paragraph{Multilingual dense retrieval and evaluation.}
Dense retrievers encode queries and passages into a shared vector space and score relevance via similarity, typically using bi-encoder architectures \citep{karpukhin2020dpr}.
Recent multilingual retrievers like Contriever~\citep{izacard2022} and BGE-M3~\citep{chen2024bgem3} are commonly benchmarked on MS~MARCO \citep{msmarco2016} and mMARCO.
These benchmarks assume that the query is in a single language (monolingual) or fully translated (cross-lingual). However, recent evidence suggests that the strong multilingual ability of LLMs in various NLP tasks does not imply robust code-mixing ability~\citep{zhang-etal-2023-multilingual}. This motivates us to study the robustness of multilingual models in code-mixed IR.
\paragraph{Mixed-language queries in IR.}
The FIRE mixed-script/mixed-language settings highlight how script variation, transliteration, and tokenization inaccuracies can affect retrieval~\citep{banerjee2016overview,cmir2016}.
This motivates our ratio-controlled study: we mix parallel query translations in embedding space to isolate language-combination effects while avoiding transliteration/tokenization noise.
Recent bilingual web-search benchmarks also report inconsistent behavior of multilingual retrievers across monolingual and mixed queries, and show that code-switched training improves robustness \citep{kim2025milqbenchmarkingirmodels}.
However, existing resources typically treat ``mixedness'' as a binary property or provide fixed sets of mixed queries, making it hard to isolate how effectiveness changes with the mixing ratio. We thus adopt a \emph{ratio-controlled} design, mixing at the embedding level to probe ratio effects while avoiding text-generation noise.

\paragraph{Code-mixed NLP benchmarks and diagnostics.}
Outside IR, code-mixed benchmark suites such as LinCE and GLUECoS report substantial pair-dependent variability and emphasize that Language Identification (LID) and segmentation errors are major sources of performance degradation \citep{lince2020,gluecos2020}.
Our controlled IR evaluation complements these diagnostics by quantifying pair and ratio-dependent effects.

\section{Method}
Our goal is to measure how multilingual dense retrieval changes as we vary the \emph{mixing ratio} between two semantically equivalent query translations in languages $(L_1, L_2)$.
We run ratio-controlled experiments under multiple document-language index settings, using \emph{embedding-level mixing} (embed-mix) as our main way to create mixed queries. We briefly explain the rationale in \ref{subsec:embedmix}.

\subsection{Task and Data}
\label{subsec:data}
We study passage ranking on mMARCO.
mMARCO provides a multilingual passage collection and multiple translations of each query aligned by a shared query ID, along with relevance judgments. Following the standard bi-encoder retrieval pipeline, we encode each passage once and index the resulting vectors using a FAISS flat index \citep{douze2024faiss}.
We L2-normalize all query and passage embeddings and use inner product scoring (equivalent to cosine similarity after normalization).
At retrieval time, we encode the query (or use a mixed query vector for embed-mix), retrieve the top $K{=}100$ passages from the index, and compute ranking metrics such as nDCG@10 on these ranked lists.

\paragraph{Setup.}
We use the 14 languages available in mMARCO:
\texttt{ar, de, en, es, fr, hi, id, it, ja, nl, pt, ru, vi, zh} and select 35 representative language pairs to support our hypotheses.
For each language pair $(L_1, L_2)$, we form a query set by matching translations with the same query ID. To keep the evaluation pool consistent across code-mixing protocols, we use a filtered subset of 1,484 sufficiently long queries. This ensures that word-level code mixing is meaningful (discussed in \S\ref{subsec:wordmix}) and avoids queries that are short and noisy. 


For our primary experiments, we use the full mMARCO corpus with 8.8 million passages. For ablations and word-mixing experiments, we use a 100k-passage subset (detailed in \S\ref{app:subset}) per document-language setting to keep computation tractable. We include all relevant passages and additional noisy documents to simulate real setups. Our main experiments use BGE-M3 \citep{chen2024bgem3}, a strong multilingual encoder for dense retrieval. We include other model families and sizes in ablation.



For each language pair $(L_1, L_2)$ we evaluate three document-language settings:
(i) monolingual $L_1$-only documents,
(ii) monolingual $L_2$-only documents,
and (iii) a union of $L_1{+}L_2$ documents (each document in 2 languages).
In every setting, the only thing that changes across conditions is the query representation (monolingual vs.\ mixed); the documents and retrieval procedure are fixed.
\begin{figure}[t]
  \centering
  \includegraphics[width=\columnwidth]{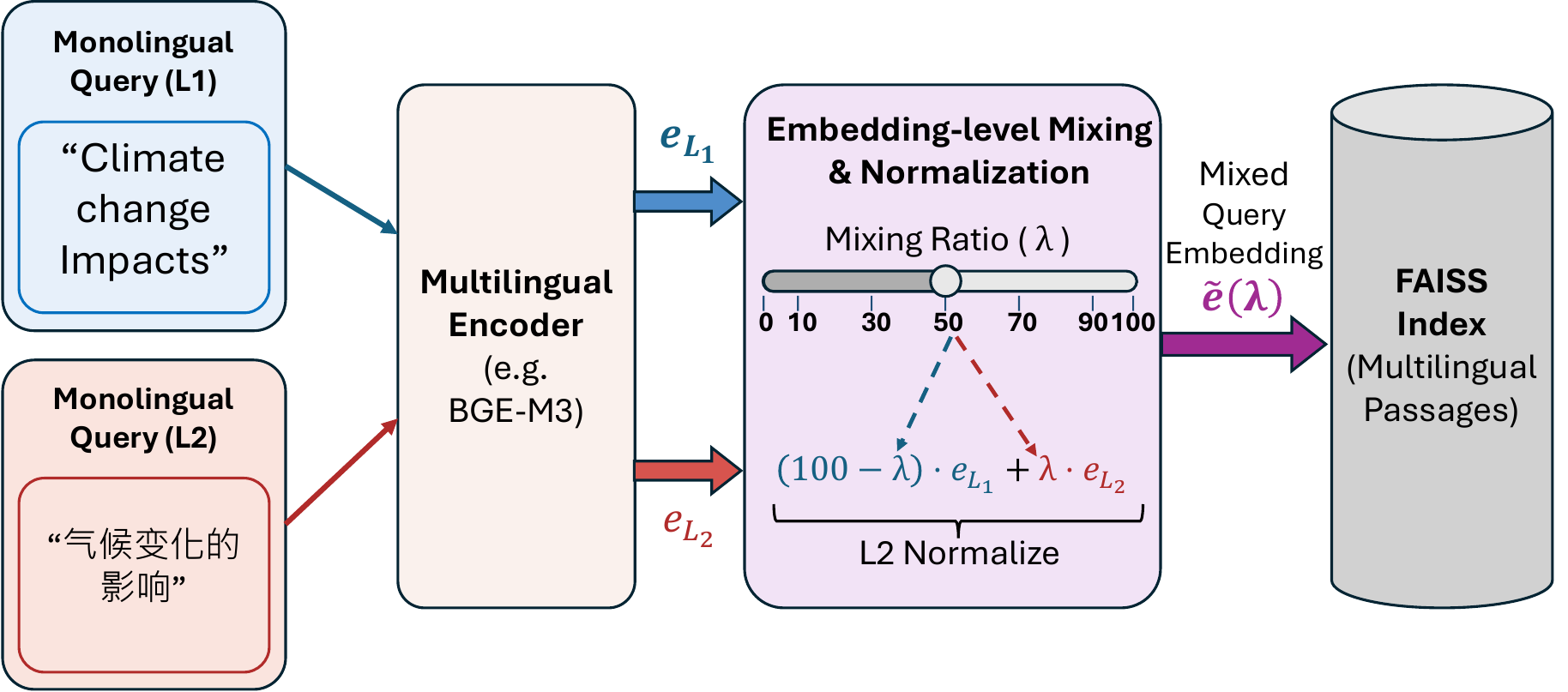}
  \caption{Illustration of Embedding-level query mixing: encode the two monolingual query translations, interpolate their embeddings with ratio $\lambda$, L2-normalize the mixed embedding, and retrieve from the index.
  }
  \label{fig:embed_mix}
  \vspace{-2mm}
\end{figure}

\subsection{Word-Level Mixing}
\label{subsec:wordmix}


We instruct an LLM to rewrite $q_{\mathrm{L_1}}$ while inserting approximately the desired number of $L_2$ content tokens (illustrated in Appendix Figure \ref{fig:word_mix}).
After generation, we estimate the realized $L_2$ share using language identification (LID) tags and discard degenerate outputs (all-$L_1$ or all-$L_2$).
To reduce sensitivity to minor ratio mismatch, we bucket accepted queries into five percentage bands (Appendix~\ref{sec:generation_details}):
$(0,20], (20,40], (40,60], (60,80], (80,100]$.
Since generation using an LLM is expensive and introduces additional variability (generation noise and tokenization/LID inaccuracies), word-mix is used only to validate embedding-level mixing.


\subsection{Embedding-Level Mixing}
\label{subsec:embedmix}

Although word-level mixing gives a natural form of code mixing, it has several problems. First, the generation process is costly: for each query, we need to instruct an LLM to generate a query of a desired mix ratio. Second, the generation quality can be poor and less controllable (Appendix \ref{subsec:word-mix-samples}): LLMs sometimes generate incomplete sentences or fail to adhere to the specified ratio. Therefore, we investigate a simpler way: directly interpolating the embeddings of two monolingual queries.  

Let $\mathbf{e}_{L_1}$ and $\mathbf{e}_{L_2}$ denote the query encoder outputs for the monolingual translations $(q_{L_1}, q_{L_2})$ under a fixed multilingual encoder.
For mixture weight $\lambda \in [0,100]$ (interpreted as the percentage of $L_2$), we define the mixed query embedding as an interpolation of the monolingual embeddings:
\begin{equation}
\tilde{\mathbf e}(\lambda) =
\frac{(1-\lambda/100)\,\mathbf e_{L_1} + (\lambda/100)\,\mathbf e_{L_2}}
     {\bigl\lVert (1-\lambda/100)\,\mathbf e_{L_1} + (\lambda/100)\,\mathbf e_{L_2} \bigr\rVert}.
\end{equation}
We iterate through $\lambda \in \{0, 10, 30, 50, 70, 90, 100\}$, 
where $\mathcal{M}=\{10,30,50,70,90\}$ denotes interior ratios and $\mathcal{E}=\{0,100\}$ denotes endpoints.
At retrieval time, we query the vector index using $\tilde{\mathbf e}(\lambda)$ directly, without using the encoder again. This flow is illustrated in Figure \ref{fig:embed_mix}.
This provides a low-cost way to study the effect of the mixing ratio on retrieval, and it also circumvents the issue of LLM generation quality and LID.

\subsection{Language Pair Metadata}
\label{subsec:langmeta}

To study how language pair affects performance, we annotate each pair with the following factors.
\textbf{Script match} indicates whether the dominant writing system matches between $L_1$ and $L_2$.
\textbf{Family distance} and \textbf{typological distance} are obtained from DistaL \citep{goot-etal-2025-distals}; unless otherwise stated, we use the continuous distances (\texttt{glot\_tree} for family and \texttt{lang2vec\_knn} for typology). 
\textbf{Resource level} follows the Microsoft linguistic diversity taxonomy \citep{joshi-etal-2020-state}, which assigns each language to resource bands 0 to 5.
As mMARCO contains no languages in bands 0 to 2, we consider band 5 as \emph{high-resource} (H) and bands~3 to 4 as \emph{lower-resource} (L), and derive pair types (H--H, H--L, L--L) accordingly.

\subsection{Evaluation and Summary Measures}
\label{subsec:eval}

\paragraph{Metrics.}
We report nDCG@10 as the primary metric, with MRR@10 and Recall@10 as secondary metrics, computed on the development split using official relevance judgments. We report these metrics in \% following \citep{kim2025milqbenchmarkingirmodels}. 

\paragraph{Measuring language-mix gains.}
For each (pair, document-language) setting, we summarize the benefit of mixing by comparing the best interior ratio (in $\mathcal{M}$) to the best endpoint (in $\mathcal{E}$). We define:
\begin{equation*}
\begin{aligned}
\texttt{endpoint\_best} = \max_{\lambda \in \mathcal{E}} \mathrm{nDCG@10}(\lambda), \\
\texttt{best\_mid}      = \max_{\lambda \in \mathcal{M}} \mathrm{nDCG@10}(\lambda), \\
\Delta                 = (\texttt{best\_mid} - \texttt{endpoint\_best}).
\end{aligned}
\label{eq:delta}
\end{equation*}
Thus $\Delta>0$ indicates that an interior mixture outperforms the best monolingual query under the same document-language setting.




\section{Results}
\label{sec:results}

We evaluate how retrieval effectiveness changes as we vary the query mixing ratio for each \emph{(language pair, document-language setting)} condition.
Unless stated otherwise, all results in this section are based on embedding-level mixing on the full mMARCO.

\subsection{Comparing Word and Embedding Mixing}
\label{sec:proxy_results}

\begin{figure*}[t]
  \centering
  \begin{subfigure}[t]{0.32\textwidth}
    \centering
    \includegraphics[width=\linewidth]{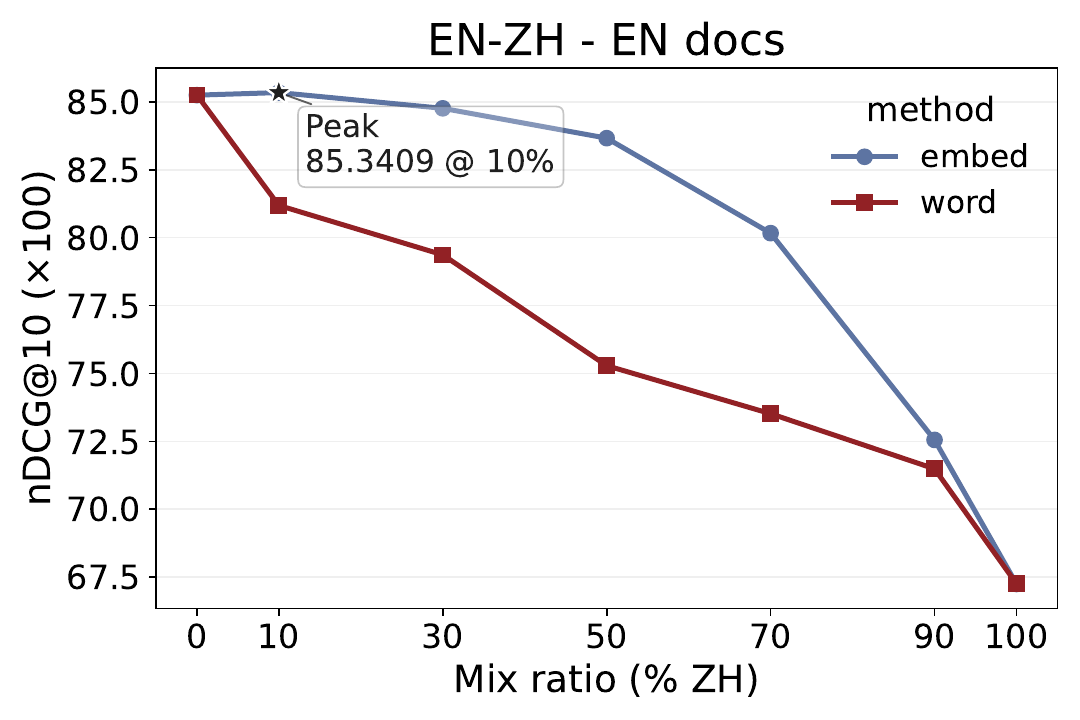}
    \caption{EN docs}
    \label{fig:enzh_proxy_en}
  \end{subfigure}\hfill
  \begin{subfigure}[t]{0.32\textwidth}
    \centering
    \includegraphics[width=\linewidth]{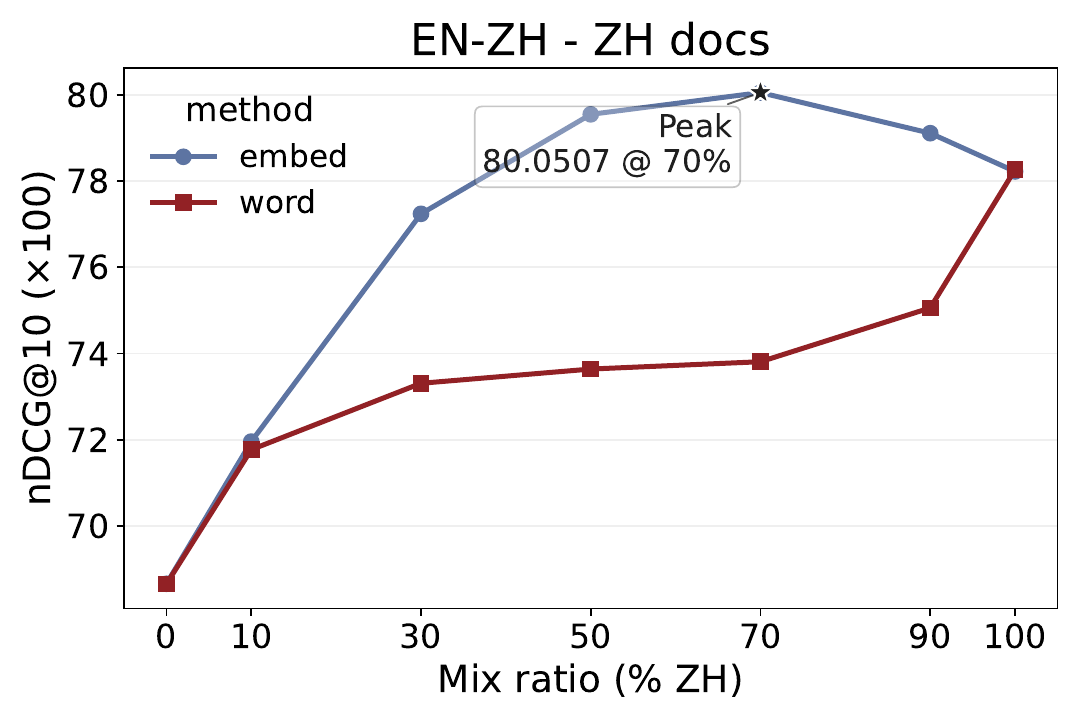}
    \caption{ZH docs}
    \label{fig:enzh_proxy_zh}
  \end{subfigure}\hfill
  \begin{subfigure}[t]{0.32\textwidth}
    \centering
    \includegraphics[width=\linewidth]{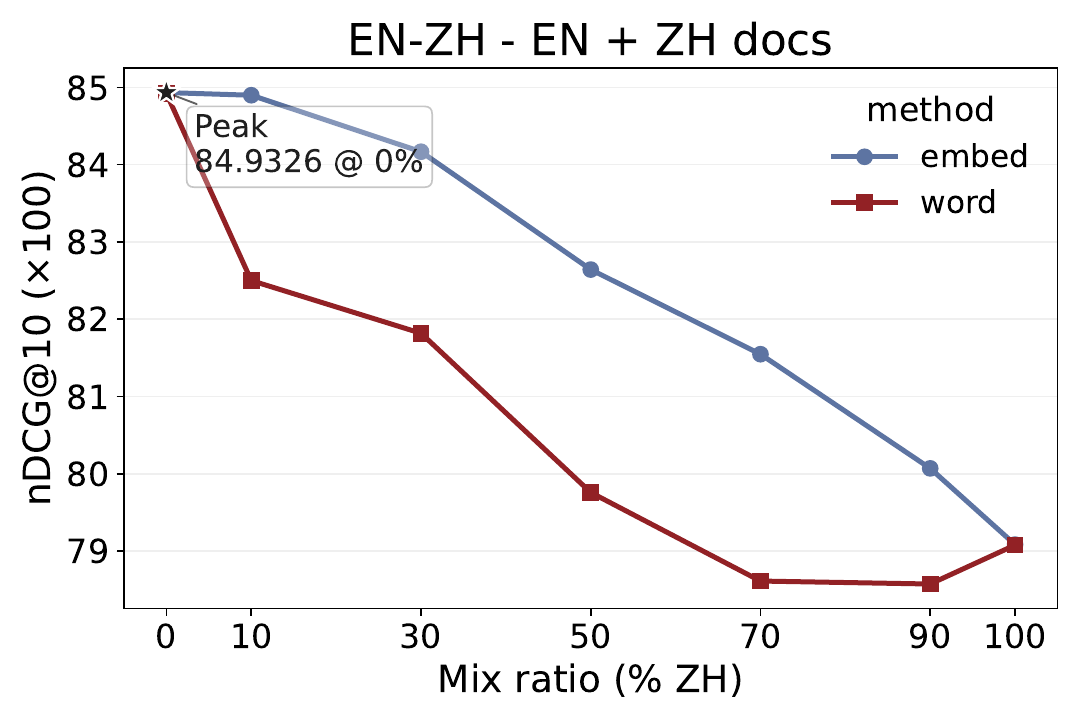}
    \caption{EN \& ZH docs}
    \label{fig:enzh_proxy_mix}
  \end{subfigure}
  \caption{Word vs. embedding-level mixing nDCG@10 on the EN--ZH pair across three document-language settings. Embed-mix preserves the overall trends observed under word-level mixing.}
  \label{fig:enzh_proxy}
\end{figure*}

We first test whether embedding-level mixing yields a consistent trend as observed when changing the mix ratio for word-level code-mixed queries.
We select the English--Chinese (EN--ZH) pair as both are high-resource languages for which an LLM should be able to perform word-level mixing, while exhibiting large differences in scripts with no shared vocabulary.
Figure~\ref{fig:enzh_proxy} compares word and embedding mixing across all three document-language settings (EN/ZH-only and EN{+}ZH).

Across all document-language settings, embed-mix matches the overall trends seen under word-mix:
(1) on EN-only documents, effectiveness decreases as the ZH proportion grows;
(2) on ZH-only documents, effectiveness increases as the ZH proportion grows;
and (3) on EN{+}ZH documents, performance smoothly interpolates between endpoints without introducing a new interior peak.
The shared trend supports using embedding-level mixing as a proxy for how performance varies with the mixing ratio in our large-scale analysis later.
We also observe that embed-mix achieves higher nDCG than word-level mixing and we attribute the absolute gap to noise specific to word-level mixing (generation variability and tokenization/LID errors), which embed-mix avoids by construction.



To further explain why embedding-level mixing tracks the ratio trends of word-level code-mixing, we analyze the geometry of word-mixed query embeddings in the encoder space.
For each aligned EN--ZH query, let $\mathbf{e}_{\mathrm{EN}}$ and $\mathbf{e}_{\mathrm{ZH}}$ be the monolingual query embeddings, and let $\mathbf{e_{\mathrm{CM}}}$ be the embedding of a word-mixed query. We define: 

\begin{equation}
\Delta=\mathbf{e}_{\mathrm{ZH}}-\mathbf{e}_{\mathrm{EN}},\qquad
d_{\mathrm{axis}}=\|\Delta\|_{2}.
\end{equation}
\begin{equation}
\mathbf{u}=\frac{\Delta}{d_{\mathrm{axis}}} \qquad p=(\mathbf{e}_{\mathrm{CM}}-\mathbf{e}_{\mathrm{EN}})^{\top}\mathbf{u}.
\end{equation}
We report two diagnostics:
\begin{equation}
r=p/d_{\mathrm{axis}},
\end{equation}
\begin{equation}
\delta=\frac{\left\|(\mathbf{e}_{\mathrm{CM}}-\mathbf{e}_{\mathrm{EN}})-p\mathbf{u}\right\|_{2}}{d_{\mathrm{axis}}}.
\end{equation}
Here, $r$ is the normalized position along the EN$\rightarrow$ZH axis in the embedding space ($r=0$ at EN, $r=1$ at ZH), and $\delta$ is the (normalized) orthogonal distance to that axis.

\begin{figure*}[t]
  \centering
  \begin{subfigure}[t]{0.49\textwidth}
    \centering
    \includegraphics[width=\linewidth]{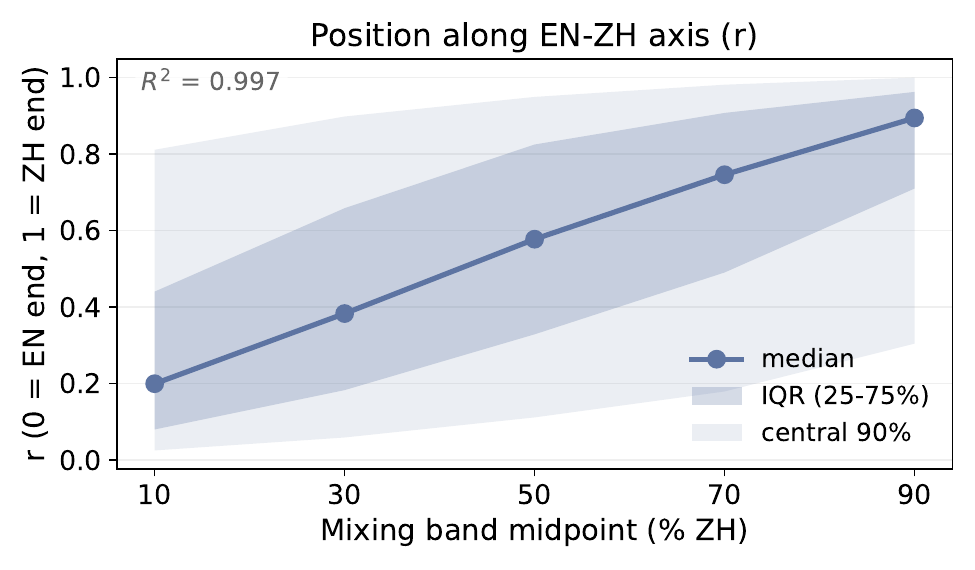}
    \caption{Axis position ($r$) vs.\ mixing band midpoint.}
    \label{fig:r_position_app}
  \end{subfigure}\hfill
  \begin{subfigure}[t]{0.49\textwidth}
    \centering
    \includegraphics[width=\linewidth]{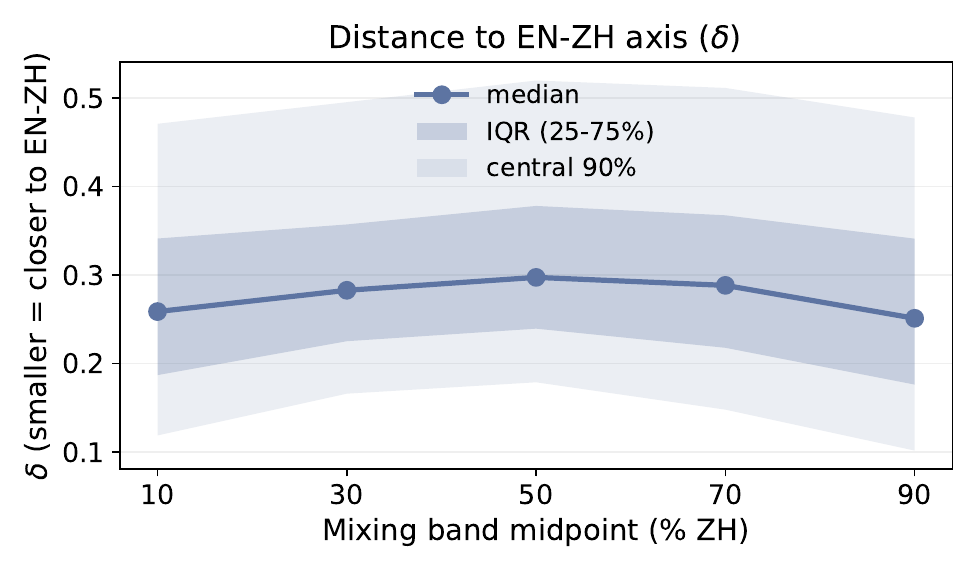}
    \caption{Off-axis distance ($\delta$) vs.\ mixing band midpoint.}
    \label{fig:delta_offset_app}
  \end{subfigure}
  \caption{\textbf{Embedding-space diagnostics (EN--ZH).} Word-mixed query embeddings move approximately along the EN$\rightarrow$ZH axis with small off-axis deviation $\delta$.}
  \label{fig:embed_panels_app}
\end{figure*}

Figure~\ref{fig:embed_panels_app} shows that as the intended ZH proportion increases, the median $r$ increases approximately linearly, while $\delta$ remains small across bands.
This indicates that word-mixed queries lie close to the interpolation trajectory between the monolingual endpoints. This partially explains why interpolating embeddings results in similar effects to word-level mixing.
Appendix~\ref{sec:wordmix_more_pairs_app} extends this validation to three additional language pairs (EN--VI, ZH--VI, HI--ID), confirming the same overarching conclusion.
Accordingly, we use embedding interpolation as the main mixing method for the remaining experiments as it is controllable, scalable, and more robust for many language pairs. 

\subsection{Global Picture: Embedding interpolation is often better than monolingual queries}
\label{subsec:global_picture}
\begin{figure*}[t]
  \centering
  \begin{subfigure}[t]{0.32\textwidth}
    \centering
    \includegraphics[width=\linewidth]{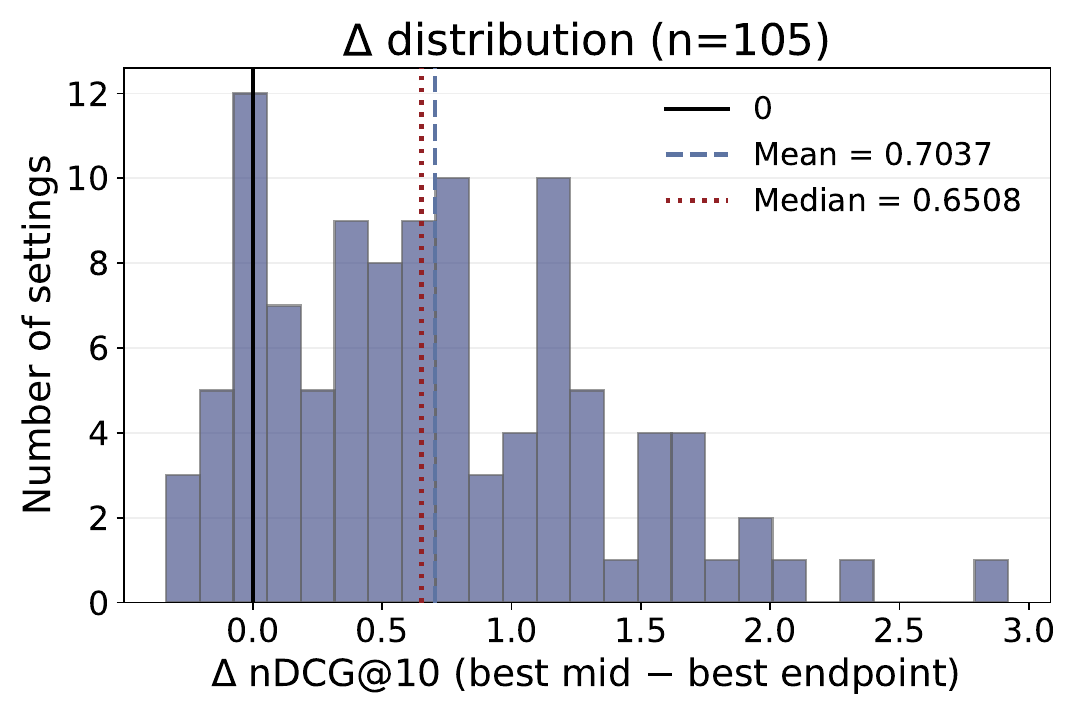}
    \caption{\textbf{Distribution of $\Delta$} over 105 settings. Embedding-level mixing is usually beneficial, and the harm is small.}
    \label{fig:delta_dist}
  \end{subfigure}\hfill
  \begin{subfigure}[t]{0.32\textwidth}
    \centering
    \includegraphics[width=\linewidth]{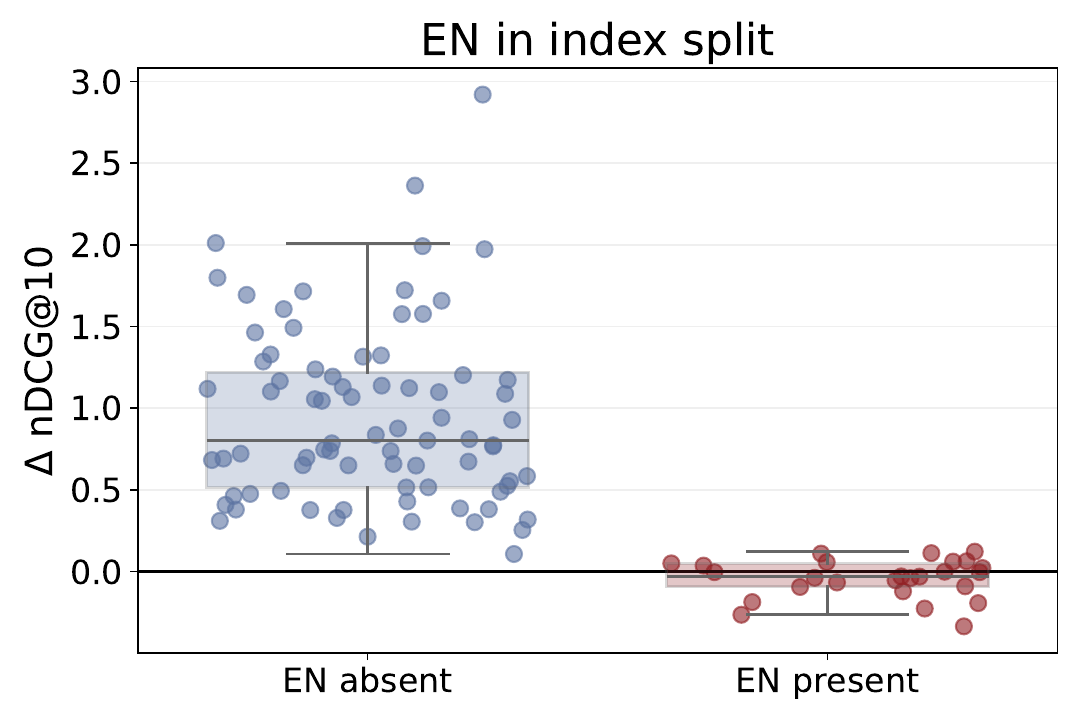}
    \caption{\textbf{Effect of English in the index.} All negative/zero cases occur when the documents include English.}
    \label{fig:en_split}
  \end{subfigure}\hfill
  \begin{subfigure}[t]{0.32\textwidth}
    \centering
    \includegraphics[width=\linewidth]{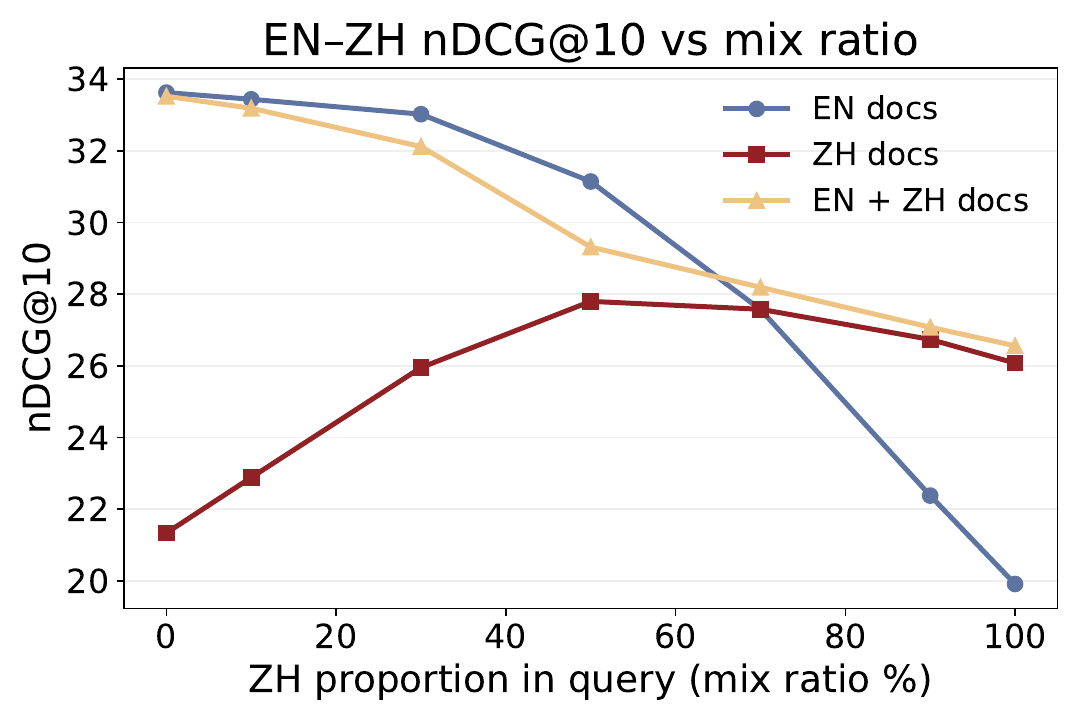}
    \caption{\textbf{EN--ZH example.} nDCG@10 vs.\ $\lambda$ (L2\%) for ZH-only, EN-only, and EN+ZH documents.}
    \label{fig:enzh_triad}
  \end{subfigure}
  \caption{Overview of effects of embedding-level code mixing.
  }
  \label{fig:mixing_overview_row}
\end{figure*}

We plot the overall distribution of the $\Delta$ (the gain of embedding-level mixing under an optimal ratio) in Figure~\ref{fig:delta_dist} for the 105 groups (35 language pairs $\times$ 3 document-language settings).
Among all groups, {88/105 (83.8\%)} have $\Delta>0$, and {17/105 (16.2\%)} have $\Delta<0$. The mean and median are $\Delta=+0.70$ and $+0.65$ respectively, and the observed range is $-0.34$ to $+2.92$. Under 95\% bootstrap intervals, 66/105 settings are \emph{reliably positive} (interval strictly above zero), 38/105 are indistinguishable from zero (interval crosses zero), and 1/105 is \emph{reliably negative}.

The largest gain occurs for {EN--AR on AR documents} ($\Delta=+2.9203$ at $\lambda^{*}=50$), while the most negative case is {EN--ZH on EN+ZH documents} ($\Delta=-0.3359$ at $\lambda^{*}=10$).
Overall, interpolating the embeddings as a way of language mixing often helps, while the worst drop is small. We hypothesize that this occurs because monolingual embeddings capture complementary semantic signals; thus, their combination yields a more robust representation. We next examine when these gains appear and what factors shape them. 



\subsection{The Strong Effects of English}

\label{subsec:en_boundary}

\paragraph{English document retrieval switches mixing effects.} Although we overall observe a positive $\Delta$ under most setups, we notice a critical factor that sharply changes the pattern: the presence of English documents. 
We split the 105 groups by whether the document-language setting includes English and plot the $\Delta$ in Figure~\ref{fig:en_split}. When the documents \emph{do not} include English (non-EN monolingual settings and non-EN+non-EN bilingual settings), all groups have $\Delta$ $>0$ with an average of $+0.95$. This suggests that there is a mixing ratio that gives better retrieval results than pure monolingual retrieval. In contrast, when English \emph{is} present in the index (EN-only or EN+X bilingual documents), all $\Delta$ are clustered near zero with a mean of $-0.04$, so mixing does not help.






\paragraph{Asymmetric effects of mixing queries with English.} Prior observations in realistic web search~\citep{kim2025milqbenchmarkingirmodels} report that \emph{intentional English mixing} can be effective for bilingual users searching \emph{English documents}, attributing the benefit primarily to increased similarity between query and document. In contrast, we find that the effect of English overrides document--query similarity. For instance, in Figure~\ref{fig:enzh_triad}, adding EN to ZH queries improves retrieval performance even in monolingual ZH documents ($\Delta=+1.72$, $\lambda^{*}=50$), but the opposite does not hold: adding ZH to EN queries consistently worsens English-document retrieval. Thus, the effect is asymmetric: mixing a non-English query with English improves performance for non-English document retrieval, but mixing English with another language does not help retrieve English documents.

\paragraph{English is the strongest mixing partner.}
\label{subsec:hub}
Beyond investigating the effect of English documents, we now investigate English as the ``mixing partner'' language for a query written in another language. We find that English is also the strongest partner language when we fix the document language and vary the partner language:
For each non-English document language $L$, English yields the largest $\Delta$ among the partners tested on $L$-only documents. We verify that this is consistent for all of the 13 languages we study. We illustrate this effect in Figure~\ref{fig:hub} and observe that English surpasses the second-best by a large margin (detailed statistics in Appendix Table~\ref{tab:en_strongest_partner}). 

We verify that these conclusions generalize to extremely low-resource languages in resource bands 0–2 in Appendix~\ref{sec:low_resource_extension_app}. While the benefits of interpolating with English stem partially from the cross-lingual preservation of entities and abbreviations (e.g., DNA, NBA), we observe that mixing improves performance even for strictly monolingual queries devoid of English loanwords (Appendix~\ref{sec:purity_filtered_app}). This indicates that the advantage of English mixing extends far beyond surface-level lexical matching. We hypothesize that this structural bias is driven by pre-training data imbalances: because English constitutes the vast majority of web corpora \citep{kudugunta2023madlad400}, it encodes the most comprehensive knowledge base. Consequently, current mainstream models develop their most robust, densely structured semantic representations within the English subspace, allowing it to act as a universal anchor that pulls mixed queries into higher-quality regions of the vector space.

\begin{figure}[bt]
  \centering
  \includegraphics[width=\linewidth]{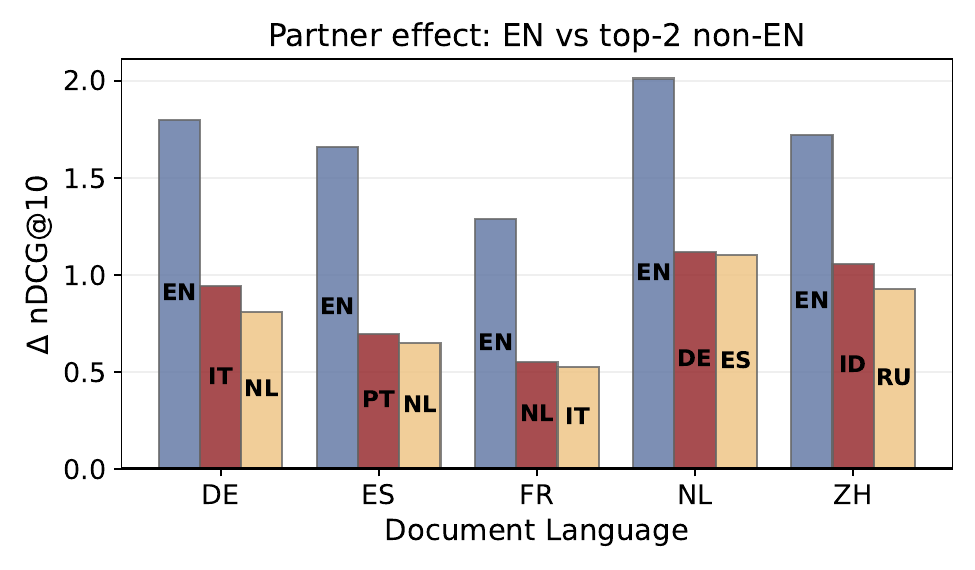}
  \caption{For multiple document languages $L$, English gives the largest $\Delta$ among tested partners on $L$-only document-language settings.}
  \label{fig:hub}
\end{figure}

\subsection{Document-Query Language Match}
\label{subsec:alignment_lambda}
\begin{figure}[bt]
  \centering
  \includegraphics[width=\linewidth]{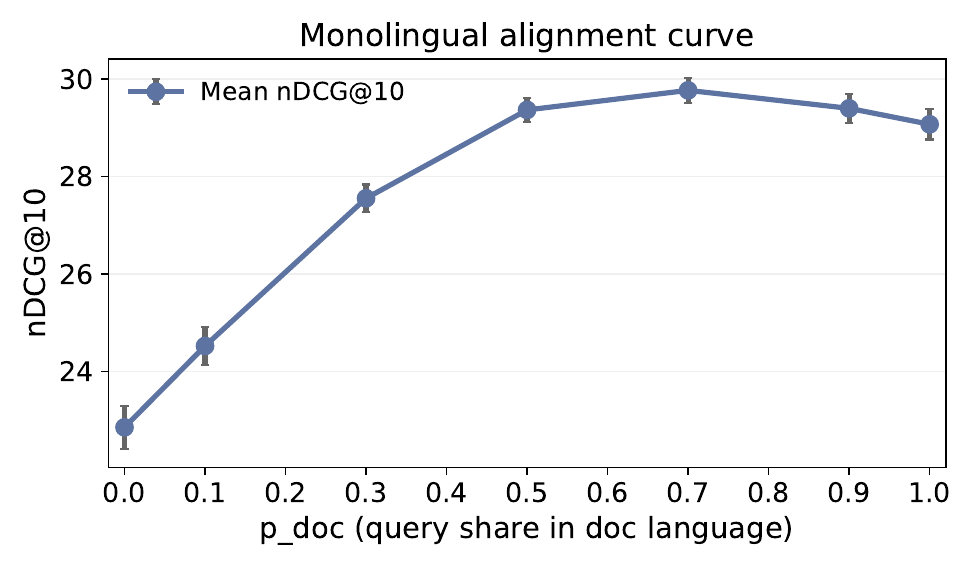}
  \caption{{Doc-language alignment in monolingual document-language settings.} We plot mean nDCG@10 as a function of $p_{\text{doc}}(\lambda)$, the fraction of the mixed query embedding coming from the indexed document language. Averaged over all monolingual document-language settings, performance increases rapidly with $p_{\text{doc}}$ and typically peaks near $p_{\text{doc}}\approx 70$.}
  \label{fig:mono_alignment}
\end{figure}

In monolingual document-language settings (single document language), effectiveness is highly related to how much of the query embedding comes from the \emph{indexed document language}.
Let $p_{\text{doc}}(\lambda)$ denote the fraction of the mixed query embedding contributed by the indexed document language (e.g., for the $L_1$-document-only setting, $p_{\text{doc}}(\lambda)=100-\lambda$). 
 Figure~\ref{fig:mono_alignment} aggregates all performance curves against $p_{\text{doc}}(\lambda)$.
 
Across all 70 monolingual document settings, the monolingual endpoint that matches the document language outperforms the opposite-language endpoint.
This is consistent with the widely observed cross-lingual alignment gap that cross-lingual retrieval is more challenging than monolingual baselines~\citep{zhang-etal-2023-miracl, conneau-etal-2020-unsupervised,litschko2023codeswitch}. Moreover, as shown in Figure~\ref{fig:mono_alignment}, as the query moves away from the document language to the other non-matching language ($p_{\text{doc}}(\lambda)$ decreases from 100 to 0), we observe an overall performance decrease. However, the performance change is not linear: mixing 50\% of query embedding in the target document language almost matches the performance of 100\% monolingual query. 
This view helps separate two effects: matching the document language remains important, but a fully monolingual query is not always optimal.
Interior mixtures can preserve enough document-language signal while adding complementary information from the other language.
Thus, the best ratio reflects a trade-off between language alignment and cross-lingual semantic enrichment.

\begin{figure}[t]
  \centering
  \includegraphics[width=\linewidth]{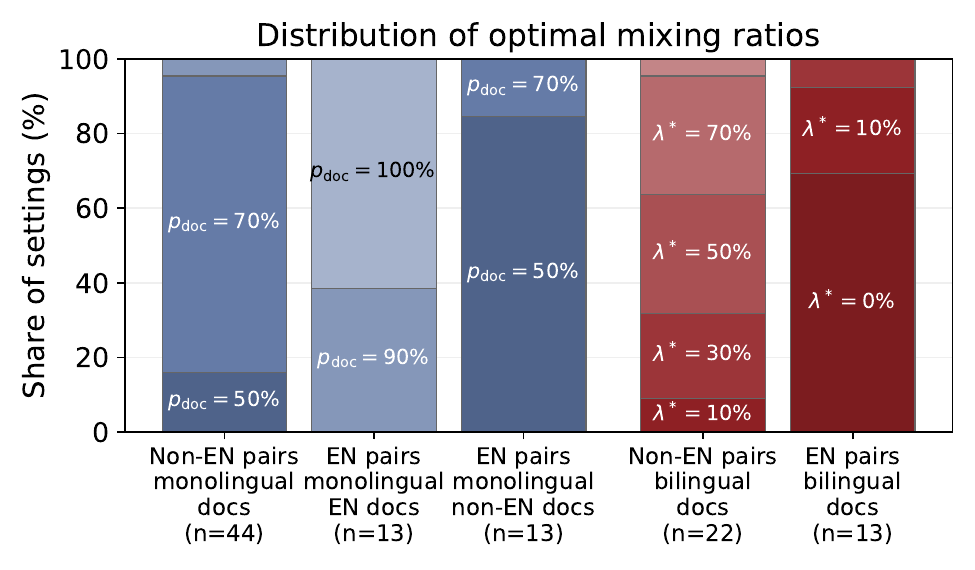}
  \caption{{Optimal query mixing ratio.} Stacked bars show where the retrieval peak across subsets: monolingual settings are labeled by $p_{\mathrm{doc}}(\lambda^*)$ and bilingual settings by $\lambda^*$ (in \%). Full results in Table \ref{tab:lambda_star_summary}.}
  \label{fig:lambda_star_summary}
\end{figure}

\paragraph{Where is the optimum $\lambda^*$?}
We now investigate what the optimal mixing ratio is for a given language pair, presented in Figure~\ref{fig:lambda_star_summary}.
Although doc-language matching endpoints are best among endpoints, the best \emph{interior} mix often occurs slightly away from the endpoint:
for non-English pairs on monolingual document-language settings, the peak is typically near $p_{\text{doc}}(\lambda^*)\approx 70$ (35/44 settings). Essentially, it means results are best when we have 70\% of the document language in the query. 
For EN pairs, optima are bimodal: in the setup with only English documents, the best is usually unmixed ($\lambda^*=0$ in 8/13), while for non-English document settings, the best often occurs near balanced mixing ($\lambda^*=50$ in 11/13). 
Finally, as shown in the rightmost two bars in Figure~\ref{fig:lambda_star_summary}, when English is absent and the documents contain both $L_1$ and $L_2$ documents (bilingual), the optimal ratio appears normally distributed due to the symmetry (i.e. we can switch $L_1$ and $L_2$). In contrast, when English is present in the pair and documents are EN+$L_2$, it is advised not to mix ($\lambda^*=0$ is the peak). These observations are consistent with our previous findings on the dominance of English.

\subsection{Other Findings}

\label{subsec:metrics}
\paragraph{Mixing affects recall more than top-rank ordering.} To show that the effects of mixing go beyond nDCG, we recompute $\Delta$ using the same definition (Eq.~\ref{eq:delta}) while replacing the retrieval metric.
Table~\ref{tab:metrics_delta} shows that nDCG@10, MRR@10, and Recall@10 broadly agree on which settings benefit from mixing, but Recall@10 exhibits the largest average improvement. In 11 settings, we observe $\Delta\text{nDCG@10}<0$ but $\Delta \text{R@10}>0$ (all within EN pairs on EN-only or EN+$L_2$ document-language settings).
This indicates that mixing can recall more relevant passages into the top-10 while slightly worsening their ordering at the top ranks.


\begin{table}[bt]
\centering
\small
\setlength{\tabcolsep}{6pt}
\begin{tabular}{lccc}
\toprule
\textbf{Metric} & \textbf{Mean $\Delta$} & \textbf{\#($\Delta>0$)} & \textbf{\#($\Delta<0$)} \\
\midrule
nDCG@10 & +0.7037 & 88 & 17 \\
MRR@10  & +0.5844 & 88 & 17 \\
R@10    & +1.2021 & 98 & 7  \\
\bottomrule
\end{tabular}
\caption{{Mixing effects across metrics.} We compute $\Delta$ per (pair, document-language) using the same endpoint-vs-interior comparison, changing only the metric.}
\label{tab:metrics_delta}
\end{table}

\begin{figure}[t]
  \centering
  \includegraphics[width=\linewidth]{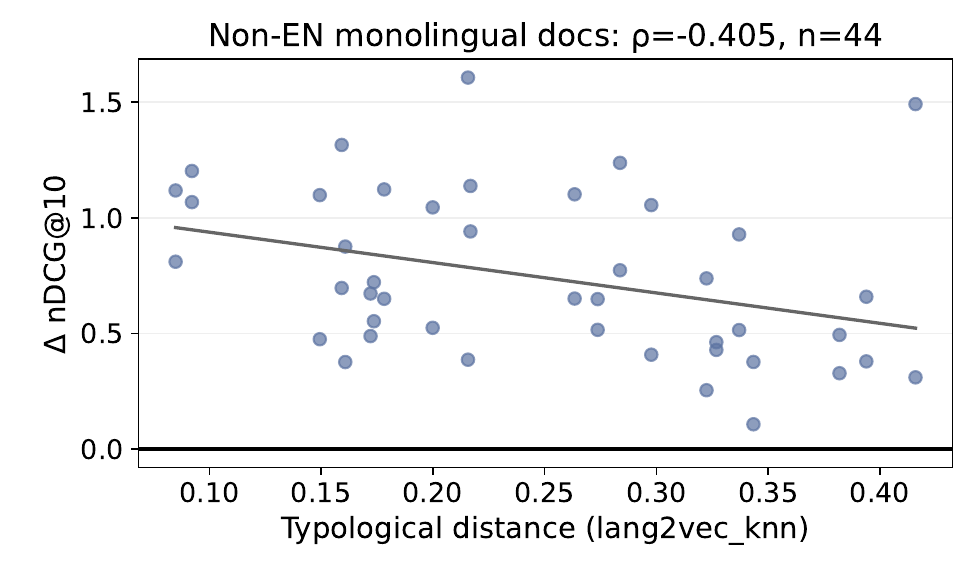}
  \caption{{Typology distance of language pairs vs.\ mixing gain $\Delta$}. The negative trend is present but modest. }
  \label{fig:typology}
\end{figure}

\label{subsec:typology}
\paragraph{Language-factor probes show weak secondary effects once English is removed.}
To study pair-intrinsic effects without the English-in-documents effect (Finding~\ref{subsec:en_boundary}), we restrict to non-English language pairs with monolingual documents. 
This yields {44 settings} (22 pairs$\times$2 document languages).
Overall, these factors are secondary compared to English-in-documents and document-language alignment (Finding~\ref{subsec:alignment_lambda}). In Figure~\ref{fig:typology}, {Typological distance} shows a moderate negative correlation with mixing gains on this subset (Spearman $\rho=-0.405$, 95\% CI [$-0.622$, $-0.124$]). Table~\ref{tab:lang_factors} summarizes the remaining three factors (script, family, resource) on the same controlled subset.
These effects are small and not cleanly monotonic given our limited language set; we therefore treat them as suggestive rather than predictive.

\begin{table}[hbt]
\centering
\scriptsize
\setlength{\tabcolsep}{3pt}
\begin{tabularx}{\linewidth}{lX}
\toprule
\textbf{Factor} & \textbf{Evidence on non-English subset} \\
\midrule
Script &
Language pairs with matching scripts benefit more from mixing ($\Delta=0.814$) than mismatched pairs ($\Delta=0.622$); the mean difference is $\Delta=0.192$ with 95\% CI [$-0.018$, $0.395$].\\ 

Family &
Family distance shows a weak negative association (Spearman $\rho=-0.306$, 95\% CI [$-0.559$, $-0.019$]); the binary family split is not strongly separative. \\
Resource &
No monotonic trend, but lower-resource pairs show larger mean gains ($\Delta=1.006$ for L--L pairs vs. $\Delta=0.560$ for H--H). \\
\bottomrule
\end{tabularx}
\caption{{The effects of script, family, and resource level on mixing gains.} Reported 95\% CIs use a cluster bootstrap over language pairs. These probes suggest only weak secondary effects. }
\label{tab:lang_factors}
\end{table}

\begin{figure}[bt]
  \centering
  \includegraphics[width=\linewidth]{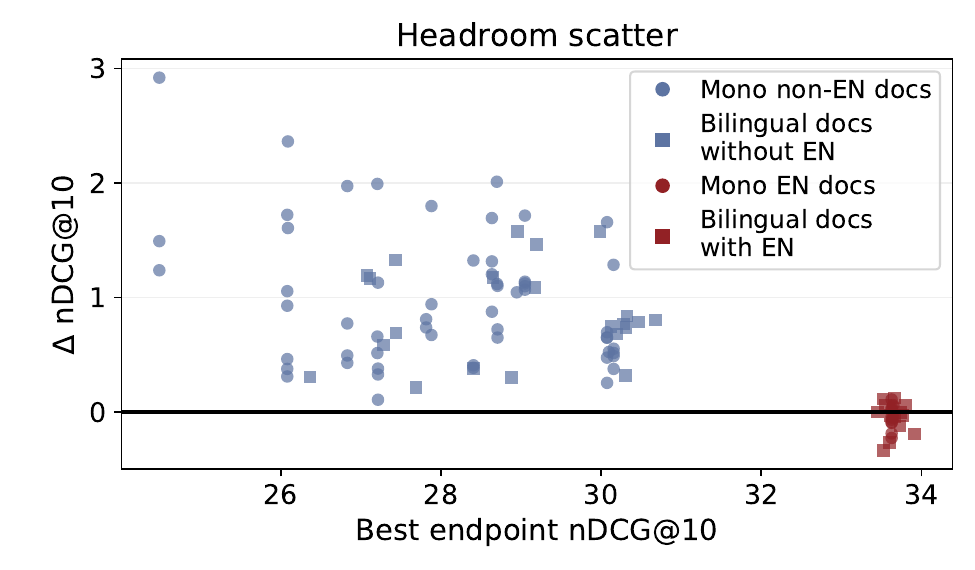}

  \caption{$\Delta$ vs.\ the stronger endpoint's nDCG@10. The negative trend is largely driven by settings with English documents, where endpoints are already strong.}
  \label{fig:headroom}
\end{figure}

\paragraph{Mixing gains also depend on how strong the best monolingual endpoint already is.}
Figure~\ref{fig:headroom} shows that $\Delta$ is negatively correlated with the better endpoint score (Spearman $\rho=-0.607$):
When a monolingual endpoint is strong, there is little room for interpolation to help.
This trend is driven by settings with English documents (EN pairs: $\rho=-0.684$), and when documents exclude English, the correlation is weak ($\rho=-0.117$).

\paragraph{Bilingual documents help mainly when English is absent.} We compare the score when retrieving from a bilingual document-language setting ($L_1{+}L_2$) against the better of the two monolingual document settings (max over $L_1$-only and $L_2$-only), holding the same query set fixed.
Table~\ref{tab:bilingual_index_gain} shows that bilingual document-language settings are usually helpful for non-English pairs, but the effect is small and inconsistent for pairs containing EN.

\begin{table}[bt]
\centering
\small
\setlength{\tabcolsep}{5pt}
\begin{tabular}{lcccc}
\toprule
\textbf{Pair type} & \textbf{Mean Gain} & \textbf{\#(Gains$>0$)} & \textbf{\#(Gains$>0.1$)} \\
\midrule
Non-EN & +0.4475 & 21/22 & 19/22 \\
EN  & +0.0101 & 8/13 & 3/13 \\
\bottomrule
\end{tabular}
\caption{Benefit of bilingual documents measured by the improvement of bilingual documents over the better monolingual document settings. Gains are common for non-EN pairs, but typically negligible for EN pairs.}
\label{tab:bilingual_index_gain}
\end{table}

\section{Ablation: Model Family and Scale}
\label{sec:ablation}
\newcommand{\maxopacity}{50} 
\newcommand{\posscale}{7.50} 
\newcommand{\negscale}{0.45} 

\newcommand{\heat}[1]{%
    \pgfmathparse{#1}%
    \ifdim\pgfmathresult pt < 0pt%
        \pgfmathparse{min(\maxopacity, pow(abs(#1) / \negscale, 0.5) * \maxopacity)}%
        \xdef\temp{\pgfmathresult}%
        \cellcolor{red!\temp!white}$#1$%
    \else%
        \pgfmathparse{min(\maxopacity, pow(#1 / \posscale, 0.5) * \maxopacity)}%
        \xdef\temp{\pgfmathresult}%
        \cellcolor{blue!\temp!white}$+#1$%
    \fi%
}

\begin{table*}[tbh]
\centering
\small
\setlength{\tabcolsep}{6pt}
\begin{tabular}{llcccccc}
\toprule
\textbf{Setting} & \textbf{Index} &
\textbf{E5} & \textbf{GTE} & \textbf{Jina} & \textbf{Qwen3 (0.6B)} & \textbf{Qwen3 (4B)} & \textbf{Qwen3 (8B)} \\
\midrule
\multicolumn{8}{l}{\textbf{English factor: English documents $\Rightarrow$ endpoint-bounded}} \\
\midrule
EN--ZH & EN docs & \heat{-0.018} (10) & \heat{-0.054} (10) & \heat{-0.147} (10) & \heat{-0.188} (10) & \heat{0.006} (30) & \heat{0.053} (10) \\
EN--AR & EN docs & \heat{-0.042} (10) & \heat{-0.217} (10) & \heat{0.010} (10) & \heat{-0.446} (10) & \heat{0.002} (10) & \heat{0.069} (10) \\
EN--DE & EN docs & \heat{0.076} (10) & \heat{-0.037} (10) & \heat{0.103} (10) & \heat{-0.019} (10) & \heat{0.157} (10) & \heat{0.053} (10) \\
\midrule
\multicolumn{8}{l}{\textbf{English factor: non-English documents $\Rightarrow$ interior gains}} \\
\midrule
EN--ZH & ZH docs & \heat{2.167} (70) & \heat{2.528} (50) & \heat{2.603} (50) & \heat{2.619} (50) & \heat{1.002} (30) & \heat{0.843} (50) \\
EN--AR & AR docs & \heat{5.684} (50) & \heat{6.088} (30) & \heat{4.331} (50) & \heat{7.384} (50) & \heat{1.789} (30) & \heat{1.190} (30) \\
EN--DE & DE docs & \heat{2.211} (50) & \heat{2.434} (30) & \heat{1.967} (30) & \heat{2.753} (50) & \heat{1.037} (30) & \heat{0.235} (30) \\
\midrule
\multicolumn{8}{l}{\textbf{EN as a strong mixing partner: doc-language fixed partner checks}} \\
\midrule
\multicolumn{8}{l}{\emph{On ZH docs: EN is strongest among tested partners}} \\
AR--ZH & ZH docs & \heat{0.468} (90) & \heat{0.116} (90) & \heat{0.608} (70) & \heat{0.454} (90) & \heat{0.528} (70) & \heat{0.702} (70) \\
EN--ZH & ZH docs & \heat{2.167} (70) & \heat{2.528} (50) & \heat{2.603} (50) & \heat{2.619} (50) & \heat{1.002} (30) & \heat{0.843} (50) \\
ID--ZH & ZH docs & \heat{1.147} (70) & \heat{1.022} (70) & \heat{1.169} (70) & \heat{0.912} (70) & \heat{0.451} (70) & \heat{0.572} (70) \\
ZH--RU & ZH docs & \heat{0.988} (30) & \heat{0.458} (30) & \heat{1.366} (30) & \heat{0.730} (30) & \heat{0.441} (30) & \heat{0.860} (30) \\
\bottomrule
\end{tabular}
\caption{{Model family and Qwen3 scale ablation.}
Each cell reports $\Delta$ nDCG@10, with $\lambda^{*}$ (percent) in parentheses. Cells are shaded by the sign and magnitude of $\Delta$ (blue: positive, red: negative; darker = larger).}
\label{tab:ablation_boundary_hub}
\end{table*}


To ensure that our findings are generalizable to other models, we experiment with alternative multilingual embedding models and different sizes of Qwen3.
Ablations are evaluated on the 100k-passage subset (Appendix~\S\ref{app:subset}). We choose the following: {multilingual-e5-large-instruct} \citep{wang2024multilingual} , {gte-multilingual-base} \citep{zhang2024mgte}, {jina-embeddings-v3} \citep{sturua2024jinaembeddingsv3multilingualembeddingstask}, {Qwen3-Embedding-0.6B/4B/8B}~\citep{qwen3embedding}.

\paragraph{Model family.} In Table \ref{tab:ablation_boundary_hub}, the English-in-index effect in Finding~\ref{subsec:en_boundary} is consistent across families: when English is indexed (EN docs), $\Delta$ is near zero, whereas on non-English documents (ZH/AR/DE docs), mixing yields clear gains.
Moreover, Table \ref{tab:ablation_boundary_hub} also validates that English is the strongest mixing partner across all four model families on ZH docs. 
The location of the best interior mixture is also stable across models:
when mixing helps on non-English-only documents, optima typically fall near balanced (often $\lambda^*=50$),
while on English documents, the best is often at the English end ($\lambda^*\in\{0,10\}$), consistent with Finding~\ref{subsec:alignment_lambda}.



\paragraph{Scale.} We further evaluate Qwen3 embedding models at three sizes, with the results in Table~\ref{tab:ablation_boundary_hub}'s rightmost columns.
The English factor (Finding~\ref{subsec:en_boundary}) persists under scaling: when English is indexed (EN docs), $\Delta$ is close to zero, whereas on non-English documents (ZH/AR/DE docs) mixing yields positive gains.
While there are occasional positive values on English-document settings, these are still much smaller compared to the gains when retrieving non-English documents, so they do not contradict our earlier conclusion that mixing does not enhance English retrieval. 
Scaling generally improves monolingual endpoint scores, which naturally reduces the magnitude of $\Delta$, but does not change most patterns.
Moreover, scale ablations also support English being the strongest mixing partner (Finding~\ref{subsec:hub}): holding document language fixed to ZH, English remains a stronger mixing partner across all model sizes. We thus conclude that the findings are generally consistent beyond a single model size or family. 

\section{Conclusion}
\label{sec:conclusion}
We presented a ratio-controlled study of mixed-language query retrieval in multilingual dense retrieval using \emph{embedding-level mixing} and discovered that interior mixtures outperform the best monolingual endpoint in many cases. Crucially, we find that these mixing effects are strongly conditioned by English. When English is absent from the indexed documents, intermediate mixing is uniformly beneficial in our tested settings; when English is present, mixing becomes neutral. We further show that, holding the document language fixed, English is the strongest mixing partner among those tested.

Overall, embedding-level mixing via interpolation provides a cheap and controlled diagnostic for {language}-mixing sensitivity. Crucially, our findings provide actionable deployment strategies for multilingual RAG practitioners: when searching English-inclusive documents, systems should default to pure English queries to avoid performance degradation. Conversely, for non-English indices, embedding mixing serves as an effective, test-time augmentation strategy to maximize retrieval recall.

For future work, extending this framework to encompass natural, organically mixed queries will be critical to addressing the sociolinguistic complexities that translation-based data cannot capture. Additionally, exploring the interpolation of multiple languages simultaneously is a promising direction for test-time scaling, potentially unlocking even greater retrieval performance across highly diverse and low-resource linguistic settings.




\FloatBarrier
\section*{Limitations}
\label{sec:limitations}
We mix two monolingual query embeddings with a chosen weight, using parallel translations of the same query.
We chose this setup because it keeps the query meaning fixed and lets us change \emph{only} the mixing weight, so the results are easier to interpret across many language pairs and index settings.
This does not cover all issues in naturally typed code-mixed queries (e.g., spelling variation, transliteration, irregular switch points, or language identification errors).
To reduce this gap, we also compare against word-level mixing for EN--ZH and add three more pairs in Appendix~\ref{sec:wordmix_more_pairs_app}. 
These checks support our ratio-trend claims, but they are still controlled generation tests, not full coverage of naturally typed mixed queries.

Our main study requires aligned queries across languages and comparable document collections, so we rely on the languages and translations provided by mMARCO.
We did not add more languages by creating new translations, because translation quality control and evaluation alignment would become a separate project, and would make it harder to keep the study consistent and reproducible.
For the same reason, we do not attempt to cover all possible language pairs: the number of pairs grows quickly.  Moreover, obtaining parallel data for low-resource languages requires substantial resources. Future work could consider expanding our study to multiple languages and more natural code-mix settings.

\section*{Ethical Considerations}
We use open-source datasets and models to study the behavior of retrieval encoders for code-mixed queries. It helps to understand the models better. We do not see any potential risk or ethical impact. 

\section*{Acknowledgments}
We appreciate the help from Dr. Barid Xi Ai, and the constructive comments by members of the NUS WING lab. Tongyao Zhu is funded by the EDB-IPP program with Sea AI Lab, Singapore. 

\bibliography{custom}
\appendix

\section{Word-mix Query Generation Details}
\label{sec:generation_details}

\begin{figure}[t]
  \centering
  \includegraphics[width=\columnwidth]{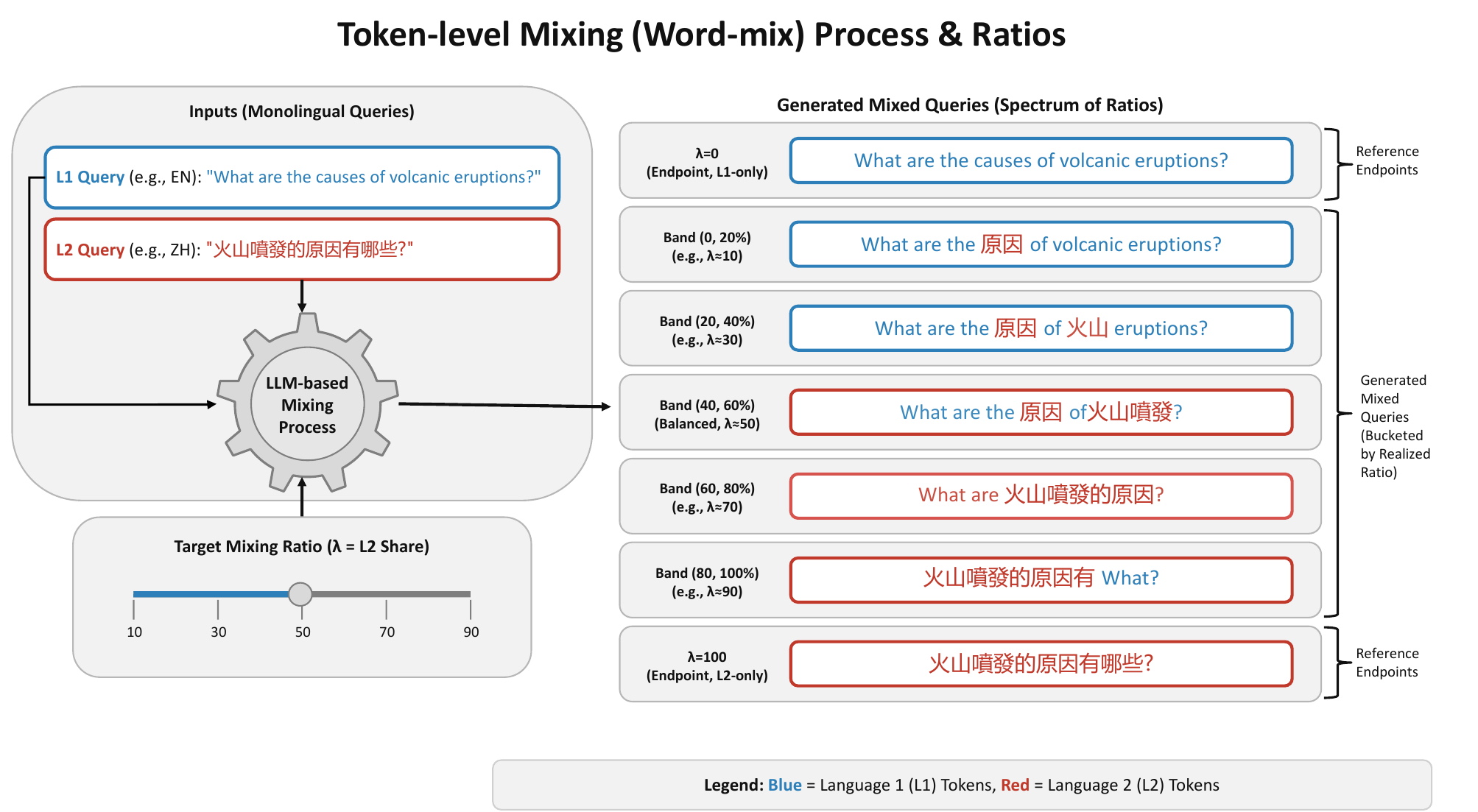}
  \caption{Illustration of Word-level query mixing via LLM, with examples.
  }
  \label{fig:word_mix}
\end{figure}

This appendix describes how we synthesize token-level code-mixed queries for the EN--ZH validation in \S\ref{subsec:wordmix}. Figure \ref{fig:word_mix} shows an overview.
The goal is \emph{not} to optimize a new retrieval method, but to sanity-check whether the trend under embedding interpolation (embed-mix) is consistent with trends observed when mixing languages at the word level.

\subsection{Pre-filtering of parallel queries}
\label{subsec:wordmix_prefilter}
Token-level mixing requires sufficient query length to meaningfully realize five ratio bands.
We therefore pre-filter the EN--ZH parallel query pairs using Stanza \citep{Qi2020stanza} tokenizer, retaining only pairs where queries in both languages have at least 6 tokens.
This removes very short queries where ratio control is ill-defined (e.g., the target bands collapse to the same few token edits).
After filtering, 2883 query pairs remain and are used as the input set for generation.

\subsection{Generation task and constraints}
For each parallel query pair $(q_{\mathrm{EN}}, q_{\mathrm{ZH}})$, we prompt a large language model (LLM) to produce one fluent, single-sentence code-mixed query for each target ZH-share band.
The prompt enforces the following high-level constraints:
(i) the output must be derived only from the given EN and ZH queries (no invented facts);
(ii) the output should not be a trivial concatenation of the two queries;
(iii) the output should avoid redundant repetition of the same meaning in two languages; and
(iv) fluency is prioritized over word-by-word alternation (reordering is allowed if it improves grammaticality).

\subsection{LLM configuration}
\label{subsec:wordmix_llm_settings}
We generate word-mix queries with \texttt{gpt-5-mini}.
We use the model's default decoding settings (we do not override sampling parameters such as temperature or top-$p$ in our API calls).
We also constrain outputs to be a single JSON object keyed by band labels, with one string value per band and no additional text.

\subsection{Operational definition of ZH share}
We estimate the realized ZH share of a generated query using Stanza tokenizer.
Let $\#\mathrm{ZH}$ and $\#\mathrm{EN}$ be the counts of tokens identified by Stanza as Chinese and English, respectively (digit-only tokens are dropped).
We define the realized ZH share as:
\[
\mathrm{ZHShare}(s) = 100 \cdot \frac{\#\mathrm{ZH}}{\#\mathrm{ZH} + \#\mathrm{EN}}.
\]
All band assignment and filtering decisions use this \emph{measured} share rather than the nominal target.

\subsection{Band targets and ratio control loop}
We use five connected ZH-share bands:
\[
(0,20],\ (20,40],\ (40,60],\ (60,80],\ (80,100).
\]
To guide the model toward each band while keeping the output length reasonable, we first count the number of English tokens in the original English query, $n_{\mathrm{EN}}$.
For each band $[L,H]$, we set an initial target at the band midpoint and convert it into a token budget:
\[
k \leftarrow \mathrm{clip}\Big(\mathrm{round}\big(\tfrac{L+H}{2}/100 \cdot n_{\mathrm{EN}}\big),\ 1,\ n_{\mathrm{EN}}-1\Big).
\]
In our implementation, we express this target as a constraint on the number of English tokens to keep:
\[
n_{\mathrm{keepEN}} = n_{\mathrm{EN}} - k,
\]
which implicitly encourages approximately $k$ Chinese tokens if the total length stays close to the original.

Generation is performed with a bounded feedback loop (small fixed number of attempts in our runs).
After each attempt, we measure the realized ZH share $\mathrm{ZHShare}(s)$:
\begin{itemize}
  \item \textbf{Accept:} if $0 < \mathrm{ZHShare}(s) < 100$ and the measured share falls inside the requested band.
  \item \textbf{Adjust and retry:} if the output is degenerate (all-EN or all-ZH) or falls outside the requested band, we adjust the token budget for that band and retry. The adjustment moves by at least one token (when possible) and scales with how far the measured share is from the band boundary.
\end{itemize}

\subsection{Filtering and band assignment}
We discard degenerate outputs that are not truly mixed (all-EN or all-ZH under the measurement above).
When an output is code-mixed but lands in a different band than requested, we \emph{assign it to the band implied by its measured share} (rather than force-editing it to hit the target band).
This reduces brittleness and avoids unnatural rewrites purely to satisfy a ratio constraint.

For band-wise retrieval comparisons, we then take the intersection of query IDs that have an accepted generation in each of the five bands.
This yields 1484 common queries, and all word-mix retrieval results are computed on this fixed set.

\subsection{Comparable query sets across bands}
To ensure band-wise retrieval comparisons are not confounded by different query subsets, we evaluate word-mix only on the intersection of query IDs that have at least one accepted output in \emph{every} band. This query subset is reduced to 1484 queries. Furthermore, to maintain consistency across all evaluations, we use this same subset of 1484 queries to evaluate all the retrieval results across all experiments in this study.

\subsection{Full generation prompt}
\label{subsec:wordmix_prompt}
The following is the exact system prompt template used in generation.
Placeholders are instantiated per query: the band labels are the five target bands, and each band is paired with a per-query constraint on the number of English tokens to keep (computed from $n_{\mathrm{EN}}$ as described above).

\begin{tcolorbox}[promptbox,title={System prompt template}]
\ttfamily\footnotesize
\setlength{\parindent}{0pt}
You are a bilingual re-writer.\par
Return a JSON object where each key is a band label and each value is ONE fluent, natural code-mixed sentence derived ONLY from the given EN \& ZH pair (reuse words/phrases; do not invent facts).\par
Code-mixing is the intra-sentence blending of two or more languages---injecting words, morphemes, or grammar from one language into an utterance in another.\par
The generated sentence should not be just a concatenation of two original sentences; you should not repeat words of the same meaning from different languages.\par
Bands to produce: <BAND\_LABELS>.\par
Fluency and Accuracy are the top priority. Preserve the original meaning fully with all information present. Avoid choppy, word-by-word alternation.\par
Ensure the Code-mixing is smooth and seamless, with good grammar and syntax in both languages.\par
You should consider to reorder or replace an English word with its Chinese counterpart (and vice-versa) to achieve best fluency.\par
Target constraints per band:\par
\hspace*{1.5em}- "<L-H>": use exactly <N\_keepEN> English words\par
\hspace*{1.5em}- "<L-H>": use exactly <N\_keepEN> English words\par
\hspace*{1.5em}- "<L-H>": use exactly <N\_keepEN> English words\par
\hspace*{1.5em}- "<L-H>": use exactly <N\_keepEN> English words\par
\hspace*{1.5em}- "<L-H>": use exactly <N\_keepEN> English words\par
Keep overall length roughly similar to the original sentence; small deviations are fine if more natural.\par
Strictly output JSON only with exactly these keys and string values. No extra commentary.\par
For example:\par
Given:\par
\hspace*{1.5em}EN: "What are the causes of volcanic eruptions?"\par
\hspace*{1.5em}ZH: "火山噴發的原因有哪些?"\par
Output:\par
\hspace*{1.5em}\{\par
\hspace*{2.5em}"0-20": "What are the 原因 of volcanic eruptions?",\par
\hspace*{2.5em}"20-40": "What are the 原因 of 火山 eruptions?",\par
\hspace*{2.5em}"40-60": "What are the 原因 of 火山噴發?",\par
\hspace*{2.5em}"60-80": "What are 火山噴發的原因?",\par
\hspace*{2.5em}"80-100": "火山噴發的原因有 what?"\par
\hspace*{1.5em}\}\par
\end{tcolorbox}

\subsection{Word-Mix Generation Samples}
\label{subsec:word-mix-samples}
Word-mix generation using an LLM is often hard to control in terms of the exact ratio and the quality of the query. Below we give some examples of the generated queries and their problems.

\paragraph{Semantic drift.}
Some generated queries drift semantically because key content words or constraints are dropped, leaving only a partial condition or an underspecified fragment. In these cases, the mixed query no longer preserves the original information need.

\begin{itemize}
  \item \textbf{qid=660957 (band 40--60):} \emph{如果 you 有 gout?} \\
  \emph{English reference (band 0):} \emph{what foods are good if you have gout?} \\
  \emph{Issue:} the main intent (\emph{``what foods are good''}) is omitted; the mixed query reduces to only the condition (\emph{``if you have gout''}).

  \item \textbf{qid=808235 (band 60--80):} \emph{什么是pill?} \\
  \emph{English reference (band 0):} \emph{what is the best erectile dysfunction pill} \\
  \emph{Issue:} the medical condition (\emph{``erectile dysfunction''}) and the superlative constraint (\emph{``best''}) are dropped; the query becomes a generic ``what is a pill?'' question.

  \item \textbf{qid=583686 (band 40--60):} \emph{内眼角痒 why?} \\
  \emph{English reference (band 0):} \emph{what cause an itch in inside corner of eye} \\
  \emph{Issue:} the query becomes a terse fragment; the causal intent is only weakly expressed by an isolated \emph{``why?''}, which makes the request less explicit than the English reference.
\end{itemize}

\paragraph{Acronym handling.}
Some queries contain acronyms, short forms, or single-letter symbols that are not naturally translated. As a result, these tokens often remain in Latin script even when the rest of the query is largely Chinese. This makes the realized Chinese-token ratio harder to control for short queries.

\begin{itemize}
  \item \textbf{qid=1082779 (band 60--80):} \emph{k在silverado中代表什么} \\
  \emph{English reference (band 0):} \emph{what does the k stand for in silverado} \\
  \emph{Issue:} \texttt{k} is a single-letter symbol that remains untranslated; in short queries, a few such tokens can noticeably shift the realized ratio.

  \item \textbf{qid=2798 (band 100):} \emph{Suddenlink 是否携带 ESPN3} \\
  \emph{English reference (band 0):} \emph{Does Suddenlink Carry ESPN3} \\
  \emph{Issue:} \texttt{ESPN3} (and the brand name \texttt{Suddenlink}) stays in English, so even the translated Chinese query contains non-Chinese tokens.

  \item \textbf{qid=257885 (band 80--100):} \emph{处理您的 SSA 退休福利申请需要多长时间？} \\
  \emph{English reference (band 0):} \emph{how long does it take to process your request for ssa retirement benefits?} \\
  \emph{Issue:} \texttt{SSA} is not translated; the acronym is also normalized to uppercase here, creating a surface mismatch with the English reference.

\end{itemize}

\paragraph{Duplication.}
Some mixed queries include redundant material from both languages. This can appear as (i) bilingual duplication of the same function words (e.g., two ways of expressing ``how'').

\begin{itemize}
  \item \textbf{qid=1099108 (band 0--20):} \emph{how does your dna fit inside your cells 如何？} \\
  \emph{English reference (band 0):} \emph{how does your dna fit inside of your cells?} \\
  \emph{Issue:} the question word is duplicated (\emph{how} + \emph{如何}), producing an unnatural mixed form.

  \item \textbf{qid=160255 (band 0--20):} \emph{Do you toast at the rehearsal dinner 吗?} \\
  \emph{English reference (band 0):} \emph{do you give a toast at rehearsal dinner} \\
  \emph{Issue:} English already marks interrogativity; adding the Chinese question particle \emph{吗} duplicates question marking and hurts fluency.

  \item \textbf{qid=904389 (band 0--20):} \emph{what time does chick-fil-a close close what 早餐?} \\
  \emph{English reference (band 0):} \emph{what time does chick-fil-a breakfast close} \\
  \emph{Issue:} \emph{close} is repeated (\emph{close close}) and an extra \emph{what} is inserted, while \emph{早餐} duplicates the already-present ``breakfast'' concept.
\end{itemize}

Additionally, to make comparisons meaningful, we have to ensure the queries are generated in all bands. If any issues arise in one band during generation, we will have to drop the entire query across bands. This makes the process highly resource-intensive if we want to ensure the quality of the code-mix generation across all mixing ratios.

\section{Constructing a 100k mMARCO subset}
\label{app:subset}

We build a fixed 100{,}000 document subset of mMARCO to reduce indexing cost while keeping document identities aligned across languages. The subset is defined purely by document IDs and is reused for every language index.

\paragraph{Step 1: include all judged-relevant documents.}
We first collect the set of unique document IDs that appear in the qrels from \texttt{BeIR/msmarco-qrels} (validation split) as relevant (i.e., documents judged relevant to at least one validation query). Every document whose ID is in this set is \emph{always} included in the subset.

\paragraph{Step 2: fill up to 100k with additional documents.}
Next, we scan through the document collection and add further documents whose IDs are \emph{not} in the qrels-derived set until the subset reaches exactly 100{,}000 documents in total. In our runs, this ``fill'' step is deterministic in the sense that we keep every encountered non-relevant document (i.e., no sampling) until the target size is reached.

\paragraph{Step 3: mirror the same subset across languages.}
mMARCO uses consistent document identifiers across languages. After selecting the 100{,}000 document IDs once, we construct each language-specific index by keeping \emph{only} those documents whose IDs belong to this fixed subset. As a result, different language indices contain different texts, but they correspond to the same underlying set of 100{,}000 documents.

\section{Additional Experiments}
\label{sec:rebuttal_appendix}

\subsection{More pairs for word-mix vs.\ embed-mix}
\label{sec:wordmix_more_pairs_app}
This study extends the proxy check in \S\ref{sec:proxy_results}.
We keep the same 100k setup, the same retriever (BGE-M3), and the same mix-ratio set $\{0,10,30,50,70,90,100\}$.
We evaluate four pairs: EN--ZH, EN--VI, ZH--VI, and HI--ID.
For each pair we test the same three document-language settings (L1 docs, L2 docs, and L1+L2 docs), giving 12 settings in total.

\begin{figure*}[t]
  \centering
  \begin{subfigure}[t]{0.32\textwidth}
    \centering
    \includegraphics[width=\linewidth]{diagrams/rebuttal_ratio_curve_ENZH_ENdocs.pdf}
    \caption{EN--ZH, EN docs}
    \label{fig:rebuttal_wordmix_enzh_en}
  \end{subfigure}\hfill
  \begin{subfigure}[t]{0.32\textwidth}
    \centering
    \includegraphics[width=\linewidth]{diagrams/rebuttal_ratio_curve_ENZH_ZHdocs.pdf}
    \caption{EN--ZH, ZH docs}
    \label{fig:rebuttal_wordmix_enzh_zh}
  \end{subfigure}\hfill
  \begin{subfigure}[t]{0.32\textwidth}
    \centering
    \includegraphics[width=\linewidth]{diagrams/rebuttal_ratio_curve_ENZH_ENplusZHdocs.pdf}
    \caption{EN--ZH, EN+ZH docs}
    \label{fig:rebuttal_wordmix_enzh_bi}
  \end{subfigure}

  \vspace{1mm}

  \begin{subfigure}[t]{0.32\textwidth}
    \centering
    \includegraphics[width=\linewidth]{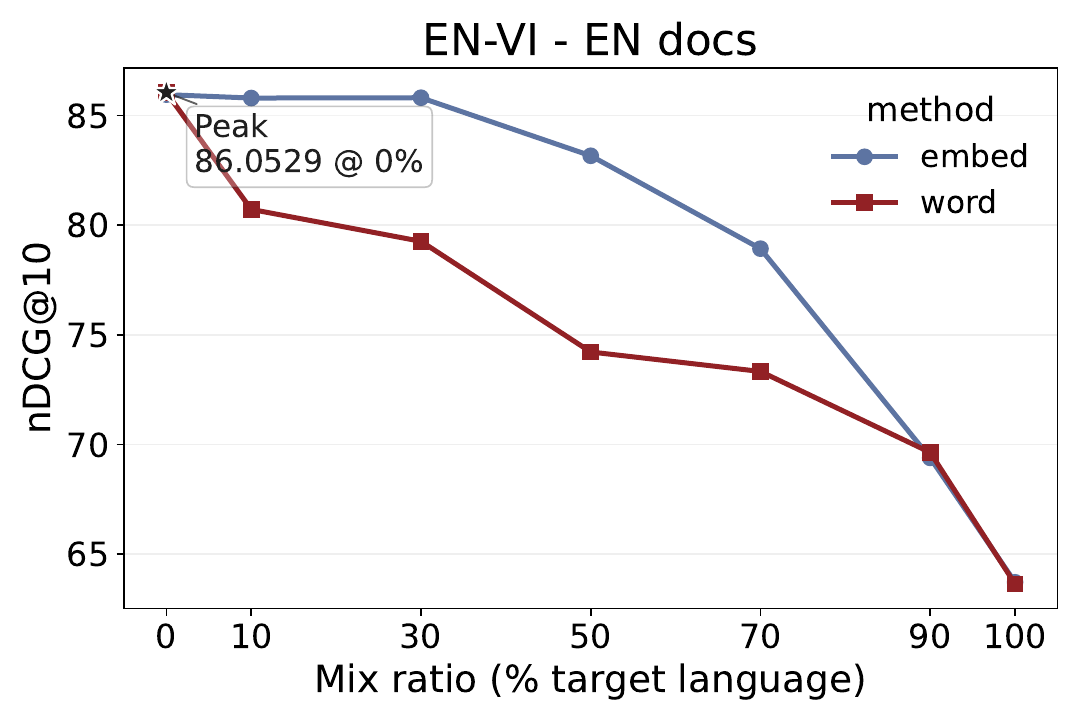}
    \caption{EN--VI, EN docs}
    \label{fig:rebuttal_wordmix_envi_en}
  \end{subfigure}\hfill
  \begin{subfigure}[t]{0.32\textwidth}
    \centering
    \includegraphics[width=\linewidth]{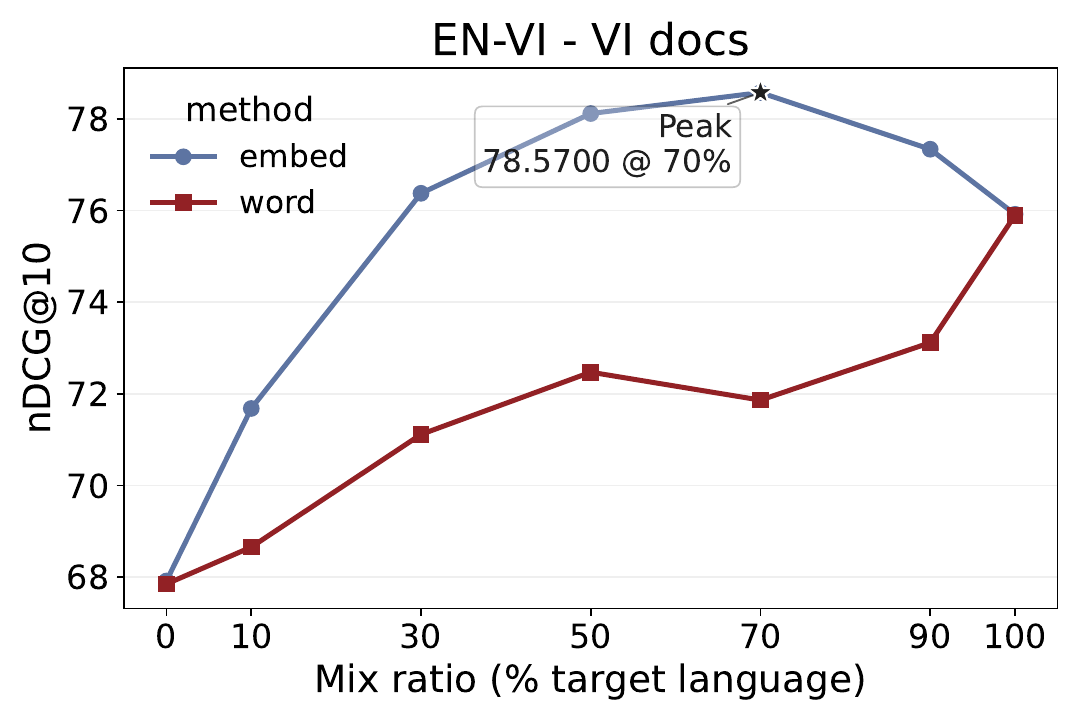}
    \caption{EN--VI, VI docs}
    \label{fig:rebuttal_wordmix_envi_vi}
  \end{subfigure}\hfill
  \begin{subfigure}[t]{0.32\textwidth}
    \centering
    \includegraphics[width=\linewidth]{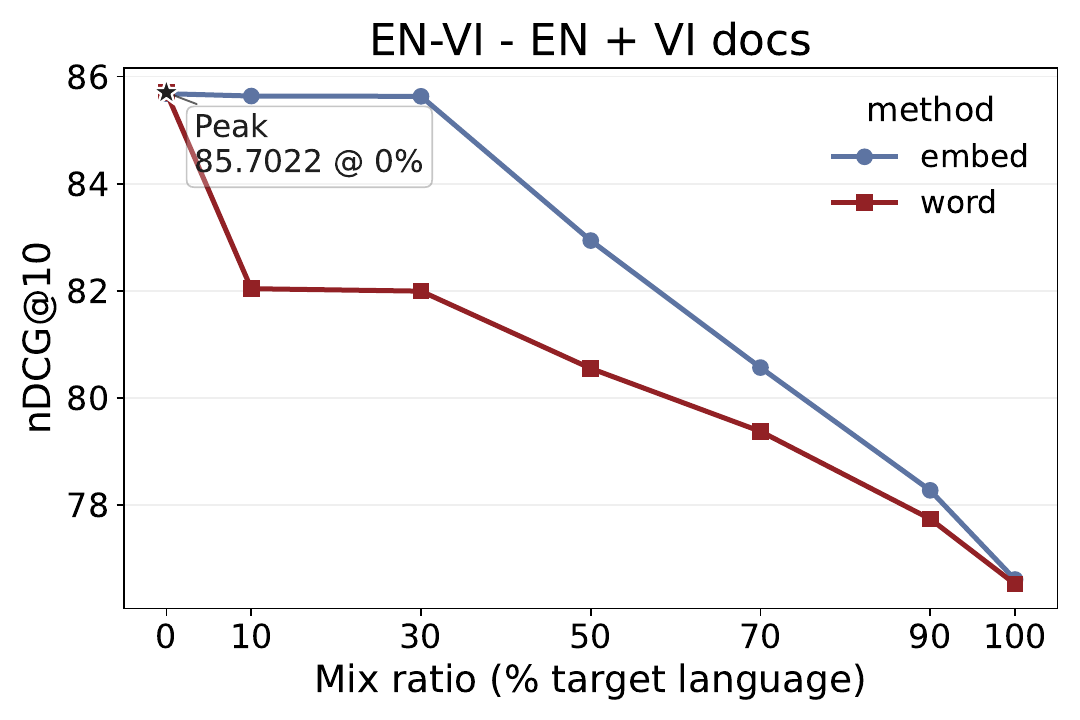}
    \caption{EN--VI, EN+VI docs}
    \label{fig:rebuttal_wordmix_envi_bi}
  \end{subfigure}
  \caption{Word vs Embed mix ratio curves for EN--ZH and EN--VI.}
  \label{fig:rebuttal_wordmix_pairs_a}
\end{figure*}
\begin{figure*}[t]
  \centering
  \begin{subfigure}[t]{0.32\textwidth}
    \centering
    \includegraphics[width=\linewidth]{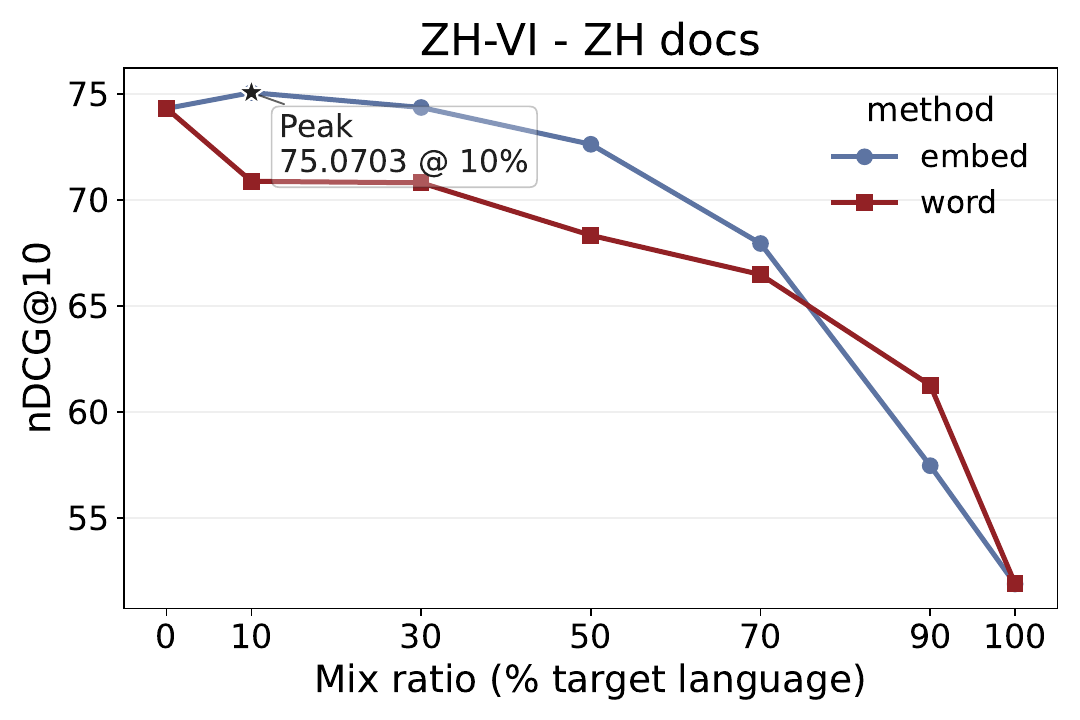}
    \caption{ZH--VI, ZH docs}
    \label{fig:rebuttal_wordmix_zhvi_zh}
  \end{subfigure}\hfill
  \begin{subfigure}[t]{0.32\textwidth}
    \centering
    \includegraphics[width=\linewidth]{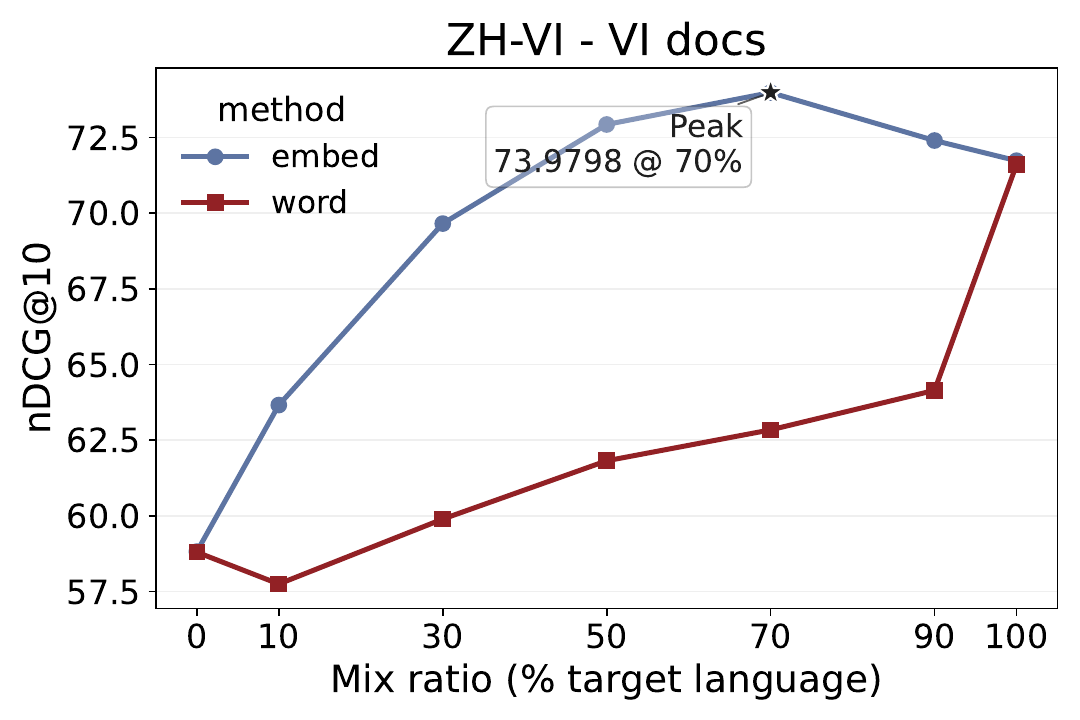}
    \caption{ZH--VI, VI docs}
    \label{fig:rebuttal_wordmix_zhvi_vi}
  \end{subfigure}\hfill
  \begin{subfigure}[t]{0.32\textwidth}
    \centering
    \includegraphics[width=\linewidth]{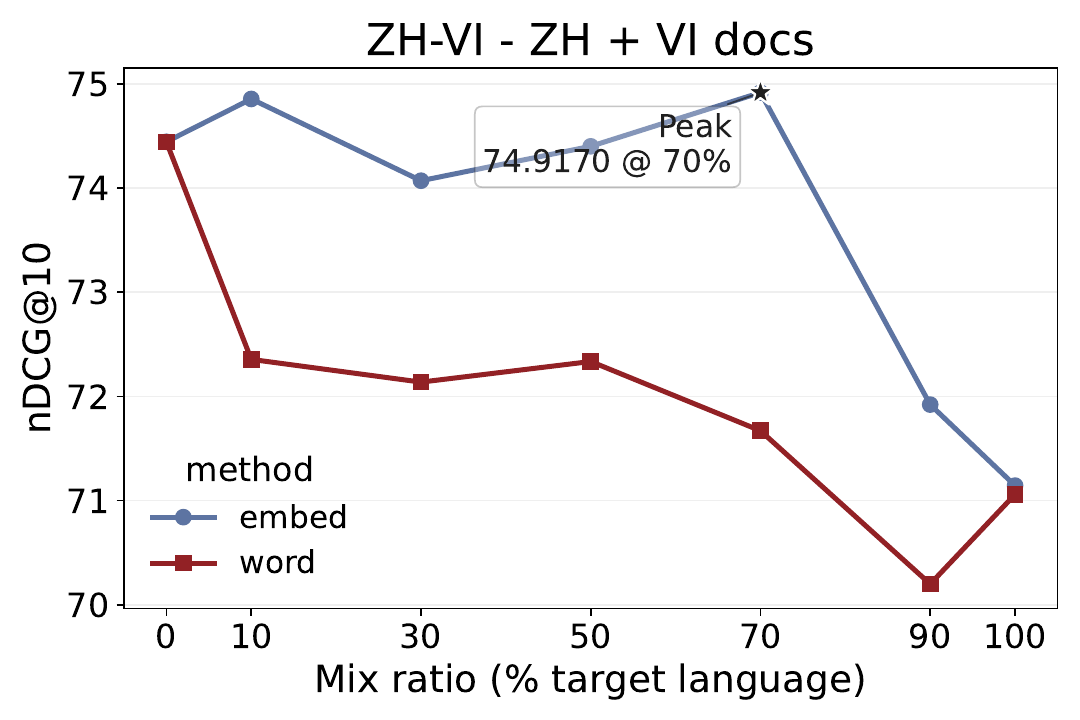}
    \caption{ZH--VI, ZH+VI docs}
    \label{fig:rebuttal_wordmix_zhvi_bi}
  \end{subfigure}

  \vspace{1mm}

  \begin{subfigure}[t]{0.32\textwidth}
    \centering
    \includegraphics[width=\linewidth]{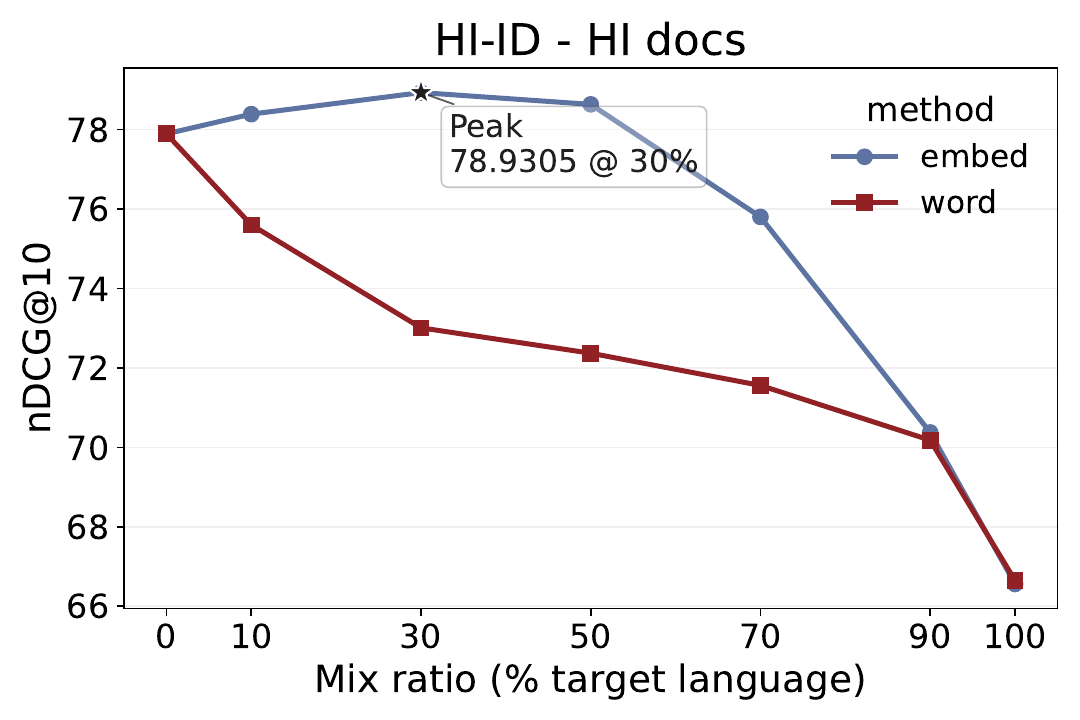}
    \caption{HI--ID, HI docs}
    \label{fig:rebuttal_wordmix_hiid_hi}
  \end{subfigure}\hfill
  \begin{subfigure}[t]{0.32\textwidth}
    \centering
    \includegraphics[width=\linewidth]{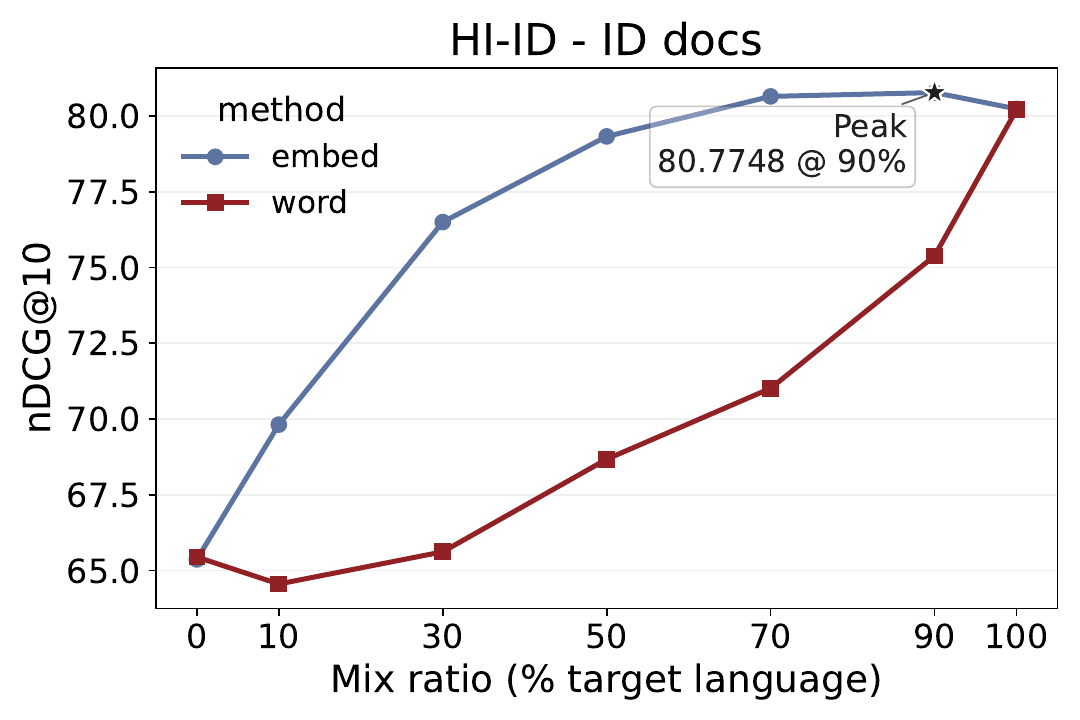}
    \caption{HI--ID, ID docs}
    \label{fig:rebuttal_wordmix_hiid_id}
  \end{subfigure}\hfill
  \begin{subfigure}[t]{0.32\textwidth}
    \centering
    \includegraphics[width=\linewidth]{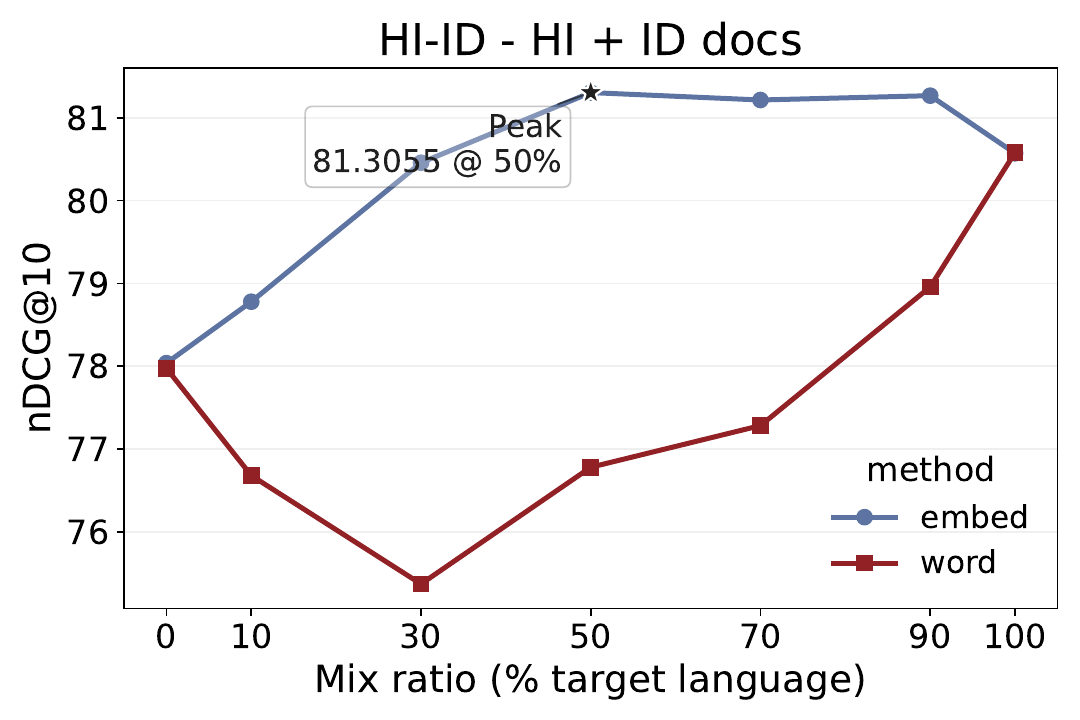}
    \caption{HI--ID, HI+ID docs}
    \label{fig:rebuttal_wordmix_hiid_bi}
  \end{subfigure}
  \caption{Word vs Embed mix ratio curves for ZH--VI and HI--ID.}
  \label{fig:rebuttal_wordmix_pairs_b}
\end{figure*}

Figures~\ref{fig:rebuttal_wordmix_pairs_a} and~\ref{fig:rebuttal_wordmix_pairs_b} show the results for each language pair.
Best word-mix retrieval is always at a pure monolingual query (0/100), while embed-mix is usually best at an interior mix ratio (10-90).
At most matching ratios (10, 30, 50, and 70), embed-mix is better than word-mix in all 12 settings, with mean gains of +3.43, +5.36, +6.09, and +5.25 nDCG@10.
This supports the same main claim as the EN--ZH-only check: the gain is not just from inserting target-language words into the query.

We reach the same conclusion as mentioned in Section~\ref{sec:proxy_results}.
First, embed-mix is a valid proxy for ratio trends beyond one pair.
Second, the gain is not explained by simple surface-level word mixing, because word-mix interior points are not the best in these settings.

\subsection{Extension to low-resource languages}
\label{sec:low_resource_extension_app}

The original mMARCO dataset contains no band 0--2 languages, so we add an extension on English paired with five low-resource languages: Sinhala (SI) in resource band 0, Nepali (NE), Khmer (KM) in band 1, Amharic (AM), and Swahili (SW) in band 2. We translated the documents to the corresponding low-resource language using the NLLB-3.3B model~\citep{nllb2022}, and translated the queries using gpt-5-mini. 
We use the same retriever (BGE-M3), the same 100k-passage setup, and the same set of ratios $\{0,10,30,50,70,90,100\}$ as in the main study.

\begin{table*}[t]
\centering
\small
\setlength{\tabcolsep}{6pt}
\begin{tabular}{lccccc}
\toprule
\textbf{Pair} & \textbf{$\Delta$ on EN docs} & \textbf{$\Delta$ on $L_2$ docs} & \textbf{$\lambda^*$ on $L_2$ docs} & \textbf{95\% CI on $L_2$ docs} & \textbf{$\Delta$ on EN+$L_2$ docs} \\
\midrule
EN--AM & +0.21 & +7.04 & 50 & [5.78, 7.90] & +0.13 \\
EN--KM & +0.13 & +2.82 & 50 & [2.09, 3.83] & -0.18 \\
EN--NE & +0.15 & +2.86 & 50 & [2.05, 3.75] & +0.13 \\
EN--SI & +0.03 & +4.60 & 50 & [3.47, 5.67] & -0.03 \\
EN--SW & +0.13 & +4.32 & 30 & [3.55, 5.12] & +0.06 \\
\bottomrule
\end{tabular}
\caption{Low-resource extension on five under-resourced languages paired with English. Each row reports the $\Delta$ (mixing gains) under the same 100k setup and ratio set as the main study.}
\label{tab:low_resource_summary}
\end{table*}

Table~\ref{tab:low_resource_summary} shows the same split as in the main study.
All five low-resource-language setups have positive $\Delta$ nDCG@10 with 95\% confidence intervals strictly above zero, with mean $\Delta=+4.33$.
The optimal ratio is balanced in four of the five cases ($\lambda^*=50$), with Swahili peaking slightly closer to English ($\lambda^*=30$).
In contrast, there are no clear gains in English-inclusive settings: the mean gain is only $+0.13$ on EN-only documents and $+0.02$ on EN+$L$ documents, and all ten 95\% intervals in these English-inclusive settings cross zero.
This suggests that the main observations persist even when we move beyond the resource range covered by mMARCO.

\subsection{Evaluating strictly monolingual documents and queries for non-Latin languages}
\label{sec:purity_filtered_app}
We next check whether the embedding-mix gains are mostly caused by impure translated text: documents or queries may still include many English words from the original dataset, as it is based on translation. To filter out impure documents and queries, we choose only the non-Latin languages: AR, HI, JA, RU, and ZH, because the classification of impure translations is more clear-cut. We classify the text as impure if the text contains any Latin-letter strings.
We then rerun the full experiments on the same settings in our main study: each language is paired with English as $L_1$ or $L_2$.
For each non-Latin language, we rebuild its index using only documents classified as pure for that language. 
For evaluation, we keep only query IDs that still have valid judged documents after document filtering, and we also remove queries classified as impure. Tables~\ref{tab:purity_doc_share} and~\ref{tab:purity_qid_retention} show the effect of such filtering. Overall, a number of documents and queries are impure, and we will investigate later what causes such language mixing. 

\begin{table}[t]
\centering
\small
\setlength{\tabcolsep}{5pt}
\begin{tabular}{lrr}
\toprule
\textbf{Language} & \textbf{Pure docs} & \textbf{Pure share (\%)} \\
\midrule
AR & 4,810,036 & 54.4 \\
HI & 6,805,767 & 77.0 \\
JA & 4,573,928 & 51.7 \\
RU & 4,964,775 & 56.2 \\
ZH & 4,180,503 & 47.3 \\
\bottomrule
\end{tabular}
\caption{Document purity check. The original dataset has 8.8 million documents. `Pure` means a document is strictly monolingual (no latin characters). }
\label{tab:purity_doc_share}
\end{table}

\begin{table}[t]
\centering
\small
\setlength{\tabcolsep}{5pt}
\begin{tabular}{lrr}
\toprule
\textbf{Document setup} & \textbf{Avg. No. Queries Kept} & \textbf{\%} \\
\midrule
L1-only docs & 892.4 & 60.1 \\
L2-only docs & 822.3 & 55.4 \\
L1+L2 docs & 724.1 & 48.8 \\
All settings & 812.9 & 54.8 \\
\bottomrule
\end{tabular}
\caption{The effect of query filtering based on purity. Baseline is 1,484 queries per setting before filtering.}
\label{tab:purity_qid_retention}
\end{table}

\begin{table}[t]
\centering
\small
\setlength{\tabcolsep}{5pt}
\begin{tabular}{p{0.44\linewidth}p{0.40\linewidth}}
\toprule
\textbf{Statistic} & \textbf{Value} \\
\midrule
$\Delta$ nDCG@10 > 0 & 37/45 (mean $\Delta$= +0.3347) \\
$\Delta$ MRR@10 > 0 & 35/45 \\
$\Delta$ Recall@10 >0 & 36/45 \\
EN in index (mean $\Delta$) & -0.0221 (6/10 negative) \\
EN not in index (mean $\Delta$) & +0.4367 (2/35 negative) \\
Largest gain & EN--HI on HI docs: +1.7161 at $\lambda^*=70$ \\
\bottomrule
\end{tabular}
\caption{Main outcomes from purity-filtered results.}
\label{tab:purity_outcomes}
\end{table}

Table~\ref{tab:purity_outcomes} shows that the conclusion remains:
mixing helps most when English is not in the document index, and helps much less when English is already in the index.
The largest gains are still on non-English monolingual indexes for EN-X pairs (EN--HI on HI docs +1.7161, EN--AR on AR docs +1.1344, EN--JA on JA docs +1.1261, EN--RU on RU docs +0.9035).

Compared with the full main study, the average gain is smaller after purity filtering (mean $\Delta=+0.3347$ here vs.\ $+0.7037$ in the full setting).
This suggests that the existence of English words in documents can partially contribute to the gain in the unfiltered corpus.


\paragraph{Investigation of translation impurity. }
To better understand what the rule-based filter is removing, we also ran an audit on the judged relevant ZH documents that had already been flagged as impure.
The taxonomy has seven classes for the Latin residue, as listed in Table~\ref{tab:zh_impurity_types}. We use ChatGPT 5.4 to assist in creating the taxonomy and running automatic data annotation.

\begin{table}[t]
\centering
\small
\setlength{\tabcolsep}{5pt}
\begin{tabular}{p{0.56\linewidth}rr}
\toprule
\textbf{Impurity label present in document} & \textbf{Docs} & \textbf{Share (\%)} \\
\midrule
English named strings & 418 & 56.4 \\
Acronyms / abbreviations & 352 & 47.5 \\
Structured codes / measures & 128 & 17.3 \\
Common English lexical items & 92 & 12.4 \\
Encoding garble & 73 & 9.9 \\
Source or template residue & 30 & 4.1 \\
Romanized non-English or Latin scientific terms & 12 & 1.6 \\
\bottomrule
\end{tabular}
\caption{Impurity labels appearing in 741 audited impure judged-relevant ZH documents. A document can contribute to multiple rows.}
\label{tab:zh_impurity_types}
\end{table}

Table~\ref{tab:zh_impurity_types} shows that the most common impurity types are English named strings (56.4\% of documents; names such as people, places, brands, or titles) and acronyms (47.5\%; short forms such as agency names or medical terms).
More than one impurity type is also common: 38.2\% of the audited documents have more than one label, and the most common pair is English named strings together with acronyms and abbreviations (15.1\% of all documents).
Structured codes/measures (such as model names, file paths, or measurement strings) also appear in 17.3\% of the documents. 
Overall, the impurity mainly comes from English entities, abbreviations or codes that are best kept in English. This suggests that the translation quality is adequate in the original dataset, and the benefit of mixing goes beyond simply lexical match. 

\subsection{Lightweight router for optimal mix-ratio guidance}
\label{sec:router_policy_app}

In this section, we investigate whether the ratio patterns observed in Section~\ref{subsec:alignment_lambda} can be operationalized into a simple routing rule. Specifically, we test a lightweight policy router that predicts the optimal mixing ratio using only the language pair and the document-language setting $(L_1, L_2, \text{Doc\_lang})$. Crucially, this router does not compute any query or document text features at inference time.

To build the router, we separate our data into a ``training'' set to build the policy and a held-out evaluation set (the same dataset used throughout the main study). For each language setting, the router evaluates a discrete set of mixing ratios $\{0, 10, 30, 50, 70, 90, 100\}$. The policy is constructed via a simple lookup table:
\begin{enumerate}
    \item Compute the mean training nDCG@10 for each ratio within a specific $(L_1, L_2, \text{Doc\_lang})$ setting.
    \item Store the ratio that yields the highest mean score for that setting.
\end{enumerate}
At runtime, the system performs a zero-cost lookup, instantly applying the stored optimal ratio for the given language setup.

\begin{table}[t]
\centering
\small
\setlength{\tabcolsep}{5pt}
\begin{tabular}{p{0.70\linewidth}p{0.20\linewidth}}
\toprule
\textbf{Held-Out Evaluation Metric} & \textbf{Value} \\
\midrule
Settings matching oracle optimal ratio & 200 / 273 \\
\midrule
Mean nDCG@10 (Router Policy) & 29.3483 \\
Mean nDCG@10 (Oracle Upper Bound) & 29.3981 \\
Mean nDCG@10 (Global 50\% Baseline) & 28.8432 \\
\midrule
Mean gain over best monolingual endpoint & +0.7190 \\
Upper-bound theoretical mean gain & +0.7634 \\
Wins / Ties / Losses vs. best monolingual & 251 / 14 / 8 \\
\bottomrule
\end{tabular}
\caption{Held-out results for the lightweight router. The ``best monolingual endpoint'' refers to the standard practice of matching the query language to the document language.}
\label{tab:router_policy_summary}
\end{table}

As shown in Table~\ref{tab:router_policy_summary}, this low-cost router successfully selects the exact optimal ratio in 200 out of 273 settings. Its overall performance (29.34) is remarkably close to the theoretical oracle upper bound (29.39), meaning it retains nearly all the potential gain over the monolingual endpoints. Furthermore, the router beats a naive baseline that applies a fixed 50\% mix ratio globally. 

These results demonstrate that complex, dynamic feature extraction is not strictly necessary for test-time query mixing. With a small amount of tuning data, practitioners can deploy a highly effective, zero-compute policy table to reliably boost multilingual retrieval performance.

\section{Additional Findings}


\begin{table}[t]
\centering
\scriptsize
\setlength{\tabcolsep}{3pt}
\begingroup
\renewcommand{\tabularxcolumn}[1]{m{#1}}
\begin{tabularx}{\linewidth}{>{\centering\arraybackslash}m{0.12\linewidth} >{\centering\arraybackslash}X >{\centering\arraybackslash}m{0.07\linewidth} >{\centering\arraybackslash}m{0.20\linewidth} >{\centering\arraybackslash}m{0.07\linewidth}}
\toprule
\multicolumn{3}{c}{\textbf{Subset}} & \multicolumn{2}{c}{\textbf{Peak}} \\
\cmidrule(lr){1-3}\cmidrule(lr){4-5}
\textbf{Pair} & \textbf{Document Index} & \textbf{\#} & \textbf{Location} & \textbf{Count} \\
\midrule
Non-EN & Monolingual; $L_1$-only or $L_2$-only & 44 &
\shortstack[c]{$p_{\text{doc}}(\lambda^*){=}0.7$\\$p_{\text{doc}}(\lambda^*){=}0.5$\\$p_{\text{doc}}(\lambda^*){=}0.9$} &
\shortstack[c]{35\\7\\2} \\
\cmidrule(lr){1-5}
with EN & Monolingual; EN-only & 13 &
\shortstack[c]{$p_{\text{doc}}(\lambda^*){=}1.0$\\$p_{\text{doc}}(\lambda^*){=}0.9$} &
\shortstack[c]{8\\5} \\
\cmidrule(lr){1-5}
with EN & Monolingual; non-EN-only & 13 &
\shortstack[c]{$p_{\text{doc}}(\lambda^*){=}0.5$\\$p_{\text{doc}}(\lambda^*){=}0.7$} &
\shortstack[c]{11\\2} \\
\cmidrule(lr){1-5}
Non-EN & Bilingual; $L_1{+}L_2$ & 22 &
\shortstack[c]{$\lambda^*{=}10$\\$\lambda^*{=}30$\\$\lambda^*{=}50$\\$\lambda^*{=}70$\\$\lambda^*{=}90$} &
\shortstack[c]{2\\5\\7\\7\\1} \\
\cmidrule(lr){1-5}
with EN & Bilingual; EN+$L_2$ & 13 &
\shortstack[c]{$\lambda^*{=}0$\\$\lambda^*{=}10$\\$\lambda^*{=}30$} &
\shortstack[c]{9\\3\\1} \\
\bottomrule
\end{tabularx}
\endgroup
\caption{{Optimal query mixing ratio}. For monolingual document-language settings, peak location is reported as $p_{\text{doc}}(\lambda^*)$; for bilingual document-language settings, peak location is reported as $\lambda^*$ (in \%).}
\label{tab:lambda_star_summary}
\end{table}

\paragraph{English as the best mixing partner.} We show the full results of mixing with different languages in Table~\ref{tab:en_strongest_partner}. We observe that across all pairs, English offers a unique advantage, showing a strong dominance in query representation.

\paragraph{Full results for the optimal query ratio.} In Table \ref{tab:lambda_star_summary} we show the full results for finding the optimal query mixing ratio under different settings, complementing figure \ref{fig:lambda_star_summary}. We see that for non-English pairs on monolingual document-language settings, the optimal mixing ratio is typically near $p_{\text{doc}}(\lambda^*)\approx 70$ (35/44 settings). 
For EN pairs, optima are bimodal: in the setup with only English documents, the best is usually unmixed ($\lambda^*=0$ in 8/13), while for non-English document settings, the best often occurs near balanced mixing ($\lambda^*=50$ in 11/13). 
Finally, when English is absent and the documents contain both $L_1$ and $L_2$ documents (bilingual), the optimal ratio appears normally distributed due to the symmetry (i.e., we can switch $L_1$ and $L_2$). In contrast, when English is present in the pair and documents are EN+$L_2$, it is advised not to mix ($\lambda^*=0$ is the peak). These observations are consistent with our previous findings on the dominance of English.

\begin{table*}[t]
\centering
\small
\setlength{\tabcolsep}{4pt}
\begin{tabularx}{\textwidth}{l c X}
\toprule
Doc language ($L$) & \# partners & Partner languages ranked by $\Delta$ on $L$-only documents \\
\midrule
AR & 3 & \textbf{EN (+2.92)} $>$ ZH (+1.49) $>$ HI (+1.24) \\
DE & 5 & \textbf{EN (+1.80)} $>$ IT (+0.94) $>$ NL (+0.81) $>$ ES (+0.74) $>$ FR (+0.67) \\
ES & 6 & \textbf{EN (+1.66)} $>$ PT (+0.70) $>$ NL (+0.65) $>$ FR (+0.65) $>$ IT (+0.47) $>$ DE (+0.25) \\
FR & 6 & \textbf{EN (+1.29)} $>$ NL (+0.55) $>$ IT (+0.52) $>$ ES (+0.52) $>$ DE (+0.49) $>$ PT (+0.38) \\
HI & 4 & \textbf{EN (+1.97)} $>$ AR (+0.77) $>$ JA (+0.49) $>$ ZH (+0.43) \\
ID & 3 & \textbf{EN (+1.32)} $>$ ZH (+0.41) $>$ VI (+0.39) \\
IT & 6 & \textbf{EN (+1.72)} $>$ DE (+1.14) $>$ NL (+1.12) $>$ ES (+1.10) $>$ PT (+1.07) $>$ FR (+1.05) \\
JA & 4 & \textbf{EN (+1.13)} $>$ RU (+0.38) $>$ HI (+0.33) $>$ ZH (+0.11) \\
NL & 5 & \textbf{EN (+2.01)} $>$ DE (+1.12) $>$ ES (+1.10) $>$ FR (+0.72) $>$ IT (+0.65) \\
PT & 4 & \textbf{EN (+1.69)} $>$ ES (+1.32) $>$ IT (+1.20) $>$ FR (+0.88) \\
RU & 3 & \textbf{EN (+1.99)} $>$ JA (+0.66) $>$ ZH (+0.51) \\
VI & 2 & \textbf{EN (+2.36)} $>$ ID (+1.61) \\
ZH & 6 & \textbf{EN (+1.72)} $>$ ID (+1.06) $>$ RU (+0.93) $>$ HI (+0.46) $>$ JA (+0.38) $>$ AR (+0.31) \\
\bottomrule
\end{tabularx}
\caption{English is the strongest mixing partner under monolingual indexing. For each non-English document language $L$, we fix the index to $L$-only documents and rank the tested partner languages by $\Delta$ (best mixed query minus the best monolingual endpoint), reported in nDCG@10 ($\%$). English attains the largest $\Delta$ for all 13 non-English document languages in our study.}
\label{tab:en_strongest_partner}
\end{table*}

\paragraph{Full Results for Model Family Ablation.} In Table~\ref{tab:appendix:ablation_family_boundary_hub}, we show the full results for model family ablation. Our findings are generally consistent across model families. In essence, the embedding models are often trained in an unsupervised way from web corpora, so we expect that the language representations are similar and English is dominant.

\begin{table*}[t]
\centering
\small
\setlength{\tabcolsep}{6pt}
\begin{tabular}{llcccc}
\toprule
\textbf{Setting} & \textbf{Index} &
\textbf{E5} & \textbf{GTE} & \textbf{Jina} & \textbf{Qwen (0.6B)} \\
\midrule
\multicolumn{6}{l}{\textbf{English factor (Finding 1): English in index $\Rightarrow$ endpoint-bounded}} \\
\midrule
EN--ZH & EN docs & $-0.0175$ (10) & $-0.0544$ (10) & $-0.1473$ (10) & $-0.1878$ (10) \\
EN--AR & EN docs & $-0.0422$ (10) & $-0.2170$ (10) & $+0.0103$ (10) & $-0.4459$ (10) \\
EN--DE & EN docs & $+0.0755$ (10) & $-0.0371$ (10) & $+0.1034$ (10) & $-0.0194$ (10) \\
\midrule
\multicolumn{6}{l}{\textbf{English factor (Finding 1): non-English index $\Rightarrow$ interior gains}} \\
\midrule
EN--ZH & ZH docs & $+2.1673$ (70) & $+2.5284$ (50) & $+2.6031$ (50) & $+2.6185$ (50) \\
EN--AR & AR docs & $+5.6839$ (50) & $+6.0879$ (30) & $+4.3305$ (50) & $+7.3837$ (50) \\
EN--DE & DE docs & $+2.2110$ (50) & $+2.4341$ (30) & $+1.9674$ (30) & $+2.7530$ (50) \\
\midrule
\multicolumn{6}{l}{\textbf{EN strongest mixing partner (Finding 2): doc-language fixed partner checks}} \\
\midrule
\multicolumn{6}{l}{\emph{On ZH docs: EN is strongest among tested partners}} \\
AR--ZH & ZH docs & $+0.4680$ (90) & $+0.1159$ (90) & $+0.6077$ (70) & $+0.4542$ (90) \\
EN--ZH & ZH docs & $+2.1673$ (70) & $+2.5284$ (50) & $+2.6031$ (50) & $+2.6185$ (50) \\
ID--ZH & ZH docs & $+1.1465$ (70) & $+1.0216$ (70) & $+1.1687$ (70) & $+0.9120$ (70) \\
ZH--RU & ZH docs & $+0.9883$ (30) & $+0.4582$ (30) & $+1.3659$ (30) & $+0.7300$ (30) \\
\midrule
\multicolumn{6}{l}{\emph{On DE docs: EN $>$ NL}} \\
DE--NL & DE docs & $+0.6357$ (30) & $+1.9691$ (50) & $+1.0221$ (30) & $+1.1582$ (30) \\
EN--DE & DE docs & $+2.2110$ (50) & $+2.4341$ (30) & $+1.9674$ (30) & $+2.7530$ (50) \\
\midrule
\multicolumn{6}{l}{\emph{On AR docs: EN $>$ ZH}} \\
AR--ZH & AR docs & $+3.0493$ (50) & $+4.4727$ (50) & $+1.8877$ (30) & $+5.7645$ (50) \\
EN--AR & AR docs & $+5.6839$ (50) & $+6.0879$ (30) & $+4.3305$ (50) & $+7.3837$ (50) \\
\midrule
\multicolumn{6}{l}{\emph{On RU docs: EN $>$ ZH}} \\
ZH--RU & RU docs & $+0.7890$ (70) & $+1.6164$ (70) & $+0.9810$ (70) & $+2.1950$ (50) \\
EN--RU & RU docs & $+2.2780$ (50) & $+2.7728$ (50) & $+3.3136$ (50) & $+3.9388$ (50) \\
\bottomrule
\end{tabular}
\caption{{Model-family ablation (100k subset index).}
Each cell reports $\Delta$ nDCG@10 (percentage points), with $\lambda^{*}$ (percent) in parentheses.
Across model families, the English-in-index boundary (Finding~\ref{subsec:en_boundary}) persists, and doc-language-fixed partner checks strengthen support for EN as the strongest mixing partner among those tested (Finding~\ref{subsec:hub}).}
\label{tab:appendix:ablation_family_boundary_hub}
\end{table*}

\section{Additional Related Work}
\paragraph{Training and augmentation with code-switching.}
Several studies explore synthetic or induced code-switching during training to improve cross-lingual robustness, sometimes with contrastive objectives that separate semantic relevance from language alignment \citep{litschko2023codeswitch,contrastivemix2024}.
While this work suggests code-mixing can be exploited as a training signal, our focus is complementary: we conduct \emph{evaluation and analysis} of off-the-shelf multilingual dense retrievers and characterize when mixing helps or hurts under a controlled setup.


\section{License of Artifacts}
The dataset we used is mMARCO, which follows the Apache-2.0 license. The BGE-M3 model uses the MIT License.

\section{Use of LLMs}
Large language models, including ChatGPT and Gemini, are used for polishing the text and fixing grammar errors. The ideation and implementation process did not involve LLMs.

\end{CJK}
\end{document}